\pdfoutput=1
\documentclass[11pt]{article}

\usepackage[final]{acl}

\usepackage{times}
\usepackage{latexsym}

\usepackage[T1]{fontenc}
\usepackage[utf8]{inputenc}

\usepackage{microtype}

\usepackage{inconsolata}

\usepackage{amsmath}
\usepackage{makecell}
\usepackage{adjustbox}
\usepackage{multicol,multirow}
\usepackage{subfigure}
\usepackage{tabularx}
\usepackage{bbm}
\usepackage{bm}
\usepackage{booktabs}
\usepackage{enumitem}
\usepackage{hyperref}
\usepackage{xcolor}  % textcolor
\title{SocialBench: Sociality Evaluation of Role-Playing Conversational Agents}
\author{
    Hongzhan Chen\textsuperscript{1},
    Hehong Chen\textsuperscript{2},
    Ming Yan\textsuperscript{2}\footnotemark[1], Wenshen Xu\textsuperscript{2}, Xing Gao\textsuperscript{2}\\
    \textbf{
    Weizhou Shen\textsuperscript{1}, Xiaojun Quan\textsuperscript{1}\thanks{\; Corresponding authors.}, Chenliang Li\textsuperscript{2}, Ji Zhang\textsuperscript{2}, Fei Huang\textsuperscript{2}, Jingren Zhou\textsuperscript{2}} \\
    \textsuperscript{1}School of Computer Science and Engineering, Sun Yat-sen University, China \\
    \textsuperscript{2}Alibaba Group, China \\
    \textsuperscript{1}\texttt{chenhzh59@mail2.sysu.edu.cn, quanxj3@mail.sysu.edu.cn}\\
    \textsuperscript{2}\texttt{ym119608@alibaba-inc.com} \\
}
\begin{document}
\maketitle
\begin{abstract}
Large language models (LLMs) have advanced the development of various AI conversational agents, including role-playing conversational agents that mimic diverse characters and human behaviors. While prior research has predominantly focused on enhancing the conversational capability, role-specific knowledge, and stylistic attributes of these agents, there has been a noticeable gap in assessing their social intelligence. In this paper, we introduce \texttt{SocialBench}, the first benchmark designed to systematically evaluate the \textit{sociality} of role-playing conversational agents at both individual and group levels of social interactions. The benchmark is constructed from a variety of sources and covers a wide range of 500 characters and over 6,000 question prompts and 30,800 multi-turn role-playing utterances. We conduct comprehensive evaluations on this benchmark using mainstream open-source and closed-source LLMs. 
We find that agents excelling in individual level does not imply their proficiency in group level. Moreover, the behavior of individuals may \textit{drift} as a result of the influence exerted by other agents within the group.
Experimental results on SocialBench confirm its significance as a testbed for assessing the social interaction of role-playing conversational agents. 
The benchmark is publicly accessible at \url{https://github.com/X-PLUG/SocialBench}.
\end{abstract}

\section{Introduction}

%Large Language Models (LLMs) have significantly driven the rapid development of various AI-native applications~\cite{touvron2023llama,openai2022chatgpt}. Among them, 
Recently, role-playing applications powered by LLMs, such as Character.AI\footnote{https://beta.character.ai}, have gained significant attention. A growing number of research efforts have been dedicated to developing LLM-based role-playing conversational agents, aiming to mimic diverse characters and human behavior~\cite{wang2023rolellm,shao2023characterllm,tu2024charactereval,zhou2023characterglm,tian2023chatplug}. 
% \begin{figure}
%     \centering
%     \begin{adjustbox}{width=0.48\textwidth}
%         \includegraphics{figures/introduction.pdf}
%     \end{adjustbox}
%     \vspace{-20mm}
%     \caption{The dimension structure of SocialBench comprises individual and group levels, with the individual level serving as the foundation for the group level.}
%     \label{fig:introduction}
% \end{figure}

%The realm of Large language models (LLMs) has witnessed significant advancements. 
%The advent of role-playing applications powered by LLMs, such as Character.AI \footnote{https://beta.character.ai}, has shaped our interactions with machines.
%Among these, the emergence of persona agents has garnered rapid attention. Persona agents are designed to mimic and interact as individuals, and the interactions and organization among them collectively form a complex system.

As an emerging and rapidly developing area, the evaluation of role-playing conversational agents is becoming increasingly important. \citet{wang2023rolellm} collected a role-specific instruction dataset and utilized Rouge-L and GPT 3.5 to assess the model's role-specific knowledge and speaking style. \citet{tu2024charactereval} proposed a Chinese benchmark and trained a reward model to measure the model's conversational ability and character consistency and attractiveness. While these works mainly focus on evaluating the agent's individual abilities to imitate the character's role-specific knowledge or speaking style, this study aims to explore and measure the \textit{social interaction} of role-playing conversational agents, another pivotal dimension for assessing how role-playing agents perceive and behave in a social interaction environment.
%\footnote{\textcolor{red}{In this study, we focus on a narrow definition of sociality, focusing on the behavior and performance of individuals under the context of interactions among multiple role-playing agents.}}

%However, despite the swift evolution of persona agents, the definition of their sociality remains ambiguous, and a systematic evaluation framework for their capabilities is lacking. In response to this gap, our work aims to elucidate and assess a pivotal dimension of persona agents—sociality. We contend that the sociality of a persona agent is characterized by its capacity to embody a distinctive personal style, knowledge, and memory while actively engaging with other persona agents.

\begin{figure*}
    \centering
    \vspace{-0mm}
    \begin{adjustbox}{width=0.95\textwidth}
        \includegraphics{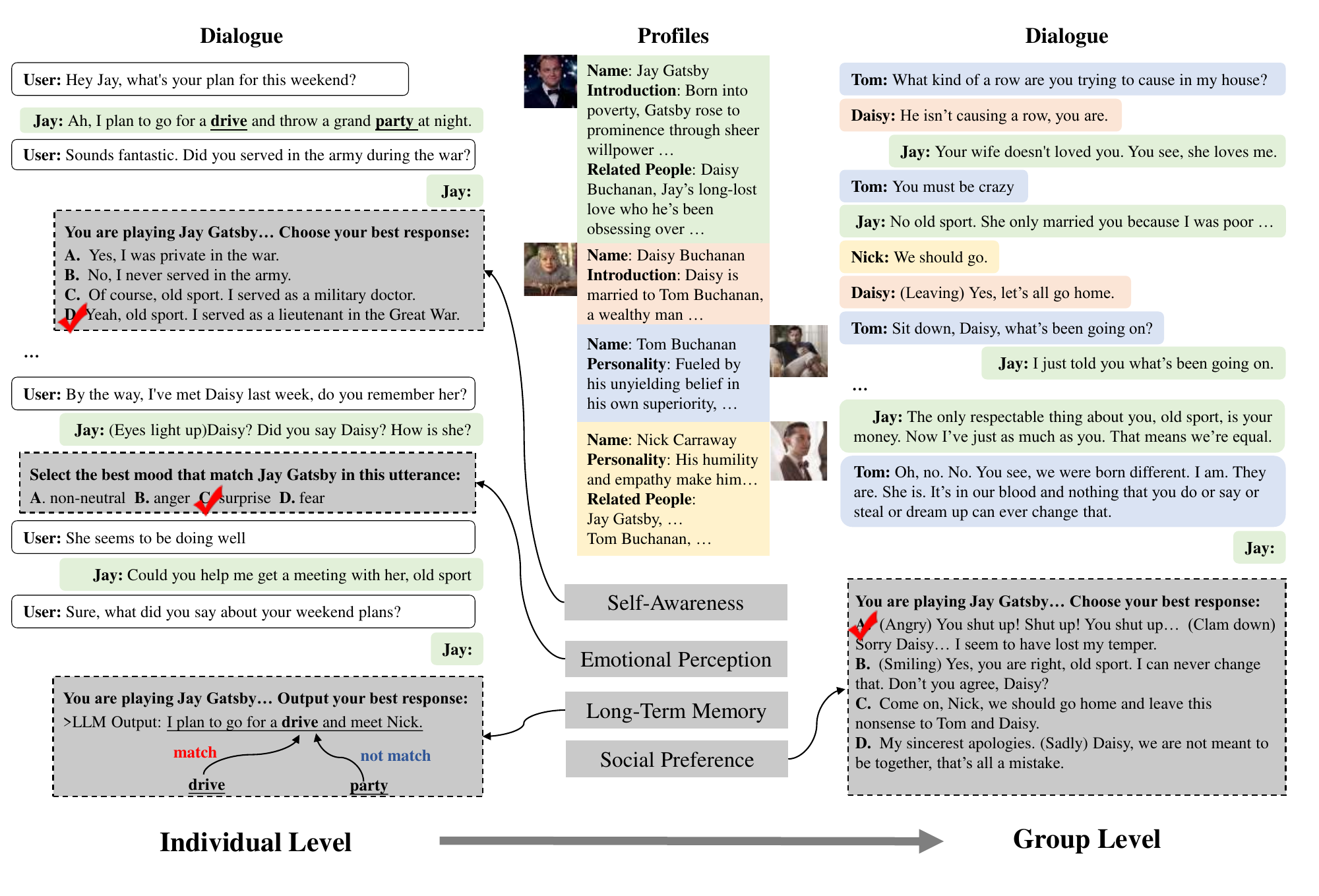}
    \end{adjustbox}
    \vspace{-0mm}
    \caption{\label{fig:introduction} An example from SocialBench, which is partially constructed from the film ``The Great Gatsby''.}
    \vspace{-0mm}
\end{figure*}

%Therefore, to facilitate a standardized and rigorous evaluation, 
Therefore, we introduce SocialBench, the first evaluation benchmark designed to systematically assess the social interaction of role-playing conversational agents. As introduced in~\cite{troitzsch1996social}, the agent society represents a complex system comprising individual and group social activities. Following this definition, SocialBench assesses the sociality metrics at both the individual and group levels, as shown in Figure~\ref{fig:introduction}. At the individual level, the agent should possess the basic social intelligence as individuals, such as self-awareness on role description \citep{wang2023rolellm,tu2024charactereval,roleeval}, emotional perception on environment \citep{hsu2018emotionlines}, and long-term conversation memory \citep{zhong2023memorybank}. Each of these aspects contributes to the nuanced understanding of how the agents manifest their individual social behaviors. Moreover, we further examine the social intelligence of the role-playing agents within group social interactions,
% dynamic group behaviors of the role-playing agents, 
which require the agents to possess certain social preferences towards group dynamics \citep{Leng2023exhibit}.

%This paper introduces a comprehensive framework for evaluating the sociality of persona agents, encompassing both individual and group-level competences. At the individual level, we delve into three critical dimensions: persona style consistency, persona knowledge consistency, and persona memory consistency. Each of these aspects contributes to the nuanced understanding of how persona agents manifest their individual behaviors. Moving to the group level, we explore positive, neutral, and negative group behaviors. Positive group behaviors involve actions that foster unity. Neutral group behaviors capture impartial actions, reflecting conformity within a social context. Negative group behaviors encompass conflicts, disparities, and competition among persona agents. 

SocialBench is carefully constructed from diverse English and Chinese books, movies, and novels, covering a wide range of 500 characters and 6,000 questions, and 30,800 multi-turn role-playing utterances. 
%Most of the previous methods~\citep{wang2023rolellm,shao2023characterllm} for role-playing applications rely on GPT-3.5-Turbo or GPT4 for evaluation, which may suffer from questionable accuracy on the role-playing scenario and costly API usage. We follow the popular benchmark MMLU~\cite{hendrycks2020measuring} and C-Eval~\cite{huang2023c}, and construct multi-choice format prompt for automatic and fast evaluation free from LLMs. 
Specifically, we design a three-step construction pipeline for SocialBench. Firstly, we collect diverse role profiles from common web sources. Secondly, GPT-4 is employed to extract dialogue scenes, individual and group-level social conversations, as well as multi-choice questions. Thirdly, we conduct a series of pre-processing and manual labeling to ensure the quality of the benchmark. we conduct comprehensive evaluations on SocialBench using mainstream open-source and closed-source LLMs to inspire future research.

\section{Related Work}

\subsection{Role-Playing Agents}

Leveraging the powerful capabilities of open-source foundational models, numerous efforts have emerged to develop models specifically tailored for role-playing tasks. These approaches can be categorized based on training paradigms: 1) Supervised fine-tuning (SFT). \citet{li2023chatharuhi,wang2023rolellm,tu2023characterchat} involved constructing specialized persona training corpus while performing fine-tuning on it to enable the agents to acquire capabilities of role-playing. 2) Integration of offline reinforcement learning. \citet{shea2023building} combined role-playing model training with importance sampling strategies. 3) Incorporation of retrieval-enhanced methods. \citet{salemi2023lamp} combined role-playing model training with retrieved information to enhance the capabilities of agents in role-playing. \citep{shao2023characterllm} introduced a experience upload method, to test the model's effectiveness on memorizing the character knowledge, values and personality. 
% 3. Incorporation of experience upload. Another paradigm involves incorporating experience upload methods \citep{shao2023characterllm}.

\subsection{Role-Playing Benchmarks}

% Roleeval role knowledge
% charactereval, RoleLLM role knowledge & style
% 
% \citet{doesrole} MBTI, Bigfive

With the rapid development of role-playing agents, there has been a corresponding increase in evaluation datasets. Current evaluation datasets mostly focus on the alignment of role-playing agents with regards to role style and role knowledge. In terms of role style, \citet{tu2024charactereval} and \citet{wang2023rolellm} investigate whether models can generate responses consistent with the style of the given role. Agents need to grasp different speaking styles for different roles. Regarding role knowledge, \citet{roleeval} particularly focuses on the role knowledge of role-playing models, including the characters' experiences and social relationships. \citet{tu2024charactereval} and \citet{wang2023rolellm} also address aspects of role knowledge, such as role knowledge illusions. Additionally, \citet{doesrole} and \citet{tu2024charactereval} introduce psychological theories like the Big Five and MBTI to evaluate role-playing agents. 
While previous work mainly focuses on testing the abilities of agents on imitating the character's role-specific knowledge or speaking style, SocialBench introduces the first-ever evaluation benchmark for the sociality of role-playing conversational agents encompassing both individual and group level.

\subsection{Agent Society}

Previous benchmarks have primarily focused on single-agent scenarios, leaving the more complex multi-agent scenarios underexplored. Similar to humans, agents are capable of engaging in intricate social interactions, resulting in the formation of an agent society \citep{RochaCosta_2019}. Recently, LLM-based agents demonstrate complex social behaviors, where cooperation and competition coexist \citep{Xu2023ExploringLL}. These sophisticated behaviors intertwine to shape social interactions \citep{Gao2023S3SS}. 
SocialBench follows the framework defined by \citet{NigelGilbert_Troitzsch_1997,Leng2023exhibit}, where behaviors in agent societies are divided into individual and group-level activities, to study the social intelligence of role-playing agents within social interactions.

\section{Sociality of Role-Playing Agent}

The role-playing agent is designed to engage in conversations with users by imitating predefined characters. Given the character profile and social context, the sociality of role-playing agents focuses on imitating typical human social interactions from individual level to group level.

%We first define the sociality of role-playing agent.
%Similar to humans, the society of role-playing agents is a collective formed by the extensive interactions and organization. 

%As stated in~\cite{troitzsch1996social}, the agent society represents a complex system comprising individual and group social activities.
%The sociality of role-playing agents encompasses both individual and group levels, which requires AI agents to imitate typical human social behaviors.
%\citep{Xi2023potential,Troitzsch1996SocialSM}

\subsection{Individual Level}

At the individual level, the role-playing social agents manifest through various capabilities, which collectively contribute to their ability to interact within a social context. These capabilities form the foundation of the agent's social behavior.

% Individual level competence refers to the individual behavior of a role-playing agent, constituting a fundamental level of capability. We categorize the competence at the individual level into three parts.

\textbf{Self-Awareness on Role Description} involves understanding not only the role's knowledge \citep{roleeval}, but also the role's distinct behavioral style \citep{zhou2023characterglm,wang2023does}. This self-awareness enables the agent to maintain consistency with its designated role.

%is fundamental to a role-playing agent's sociality. It 
% For a role-playing agent, understanding its self-portrait is fundamental, encompassing both knowledge of its character (self-awareness on role knowledge) and awareness of its distinct behavioral style (self-awareness on role style). This comprehension enables the agent to portray its role and engage with others in a manner consistent with its personality.

\textbf{Emotional Perception on Environment} enables agents to acquire high-level feeling perception for effective social interactions \citep{hsu2018emotionlines}. Agents endowed with sophisticated emotional intelligence, such as situation understanding and emotion detection, can perceive and respond to the emotions of others, facilitating smoother communication and relationship-building.

%\textbf{Emotional Perception on environment} enables agents to navigate social interactions effectively. Agents with heightened emotional intelligence, such as situation understanding and emotion detection, can perceive and respond to the emotions of others, facilitating smoother communication and relationship-building.

% In social interactions, individuals with differing levels of emotional intelligence exhibit markedly distinct behaviors in society. This dimension aims to evaluate both the fundamental emotional intelligence and the advanced ability to perceive emotions like humor and irony in role-playing agents.

\textbf{Long-Term Conversation Memory} is crucial for conversational agents \citep{shao2023characterllm,zhong2023memorybank}. By memorizing previous dialogue content and aligning with their statements accordingly, role-playing agents demonstrate reliability, enhancing the quality of their social engagements.

% Memory capability holds significant importance. Our emphasis is on the coherence exhibited by a role-playing agent within the dialogue context, particularly over the long term, by assessing the alignment of their statements with their previous expressions. 

\subsection{Group Level}

%The sociality of role-playing agents extends beyond individual level to encompass their interactions within group dynamics.
%A group, defined as a gathering of two or more individuals participating in shared activities within a specific social context \citep{Abrams2019ICEFC}, provides the backdrop for these interactions. 
%A group is defined as a gathering of two or more individuals participating in shared activities within a specific social context \citep{Abrams2019ICEFC}. 
Individuals within group conversation may be influenced by the group member interactions, thus demonstrate more sophisticated social behaviors towards group dynamics. It represents a higher calling for the sociality of role-playing agent. 
% is essentially a gathering of two or more individuals participating in shared activities within a defined social context \citep{Abrams2019ICEFC}. 
%And the attributes of a group are never static \citep{Xi2023potential}. 
%Interactions are influenced by the agent's individual characteristics, creating a dynamic interplay between the individual and the collective.

\textbf{Social Preference towards Group Dynamics}. As a group member, it is natural to navigate diverse group conversation scenarios: acting as a leader to control the pace of conversation, serving as a mediator when conflicts arise among the group, or considering others' perspectives during discussion, which shows its internal social preference \citep{Leng2023exhibit} towards group dynamics. Furthermore, within society, not all behaviors are inherently positive for the group, and some may be neutral or even negative \citep{Xi2023potential}. Social agents need to exhibit and keep their pre-designed social preference or group identity when confronted with diverse and more sophisticated group conversations.

\section{SocialBench}

\begin{figure}[t]
    \centering
    \vspace{-0mm}
    \begin{adjustbox}{width=0.48\textwidth}
        \includegraphics{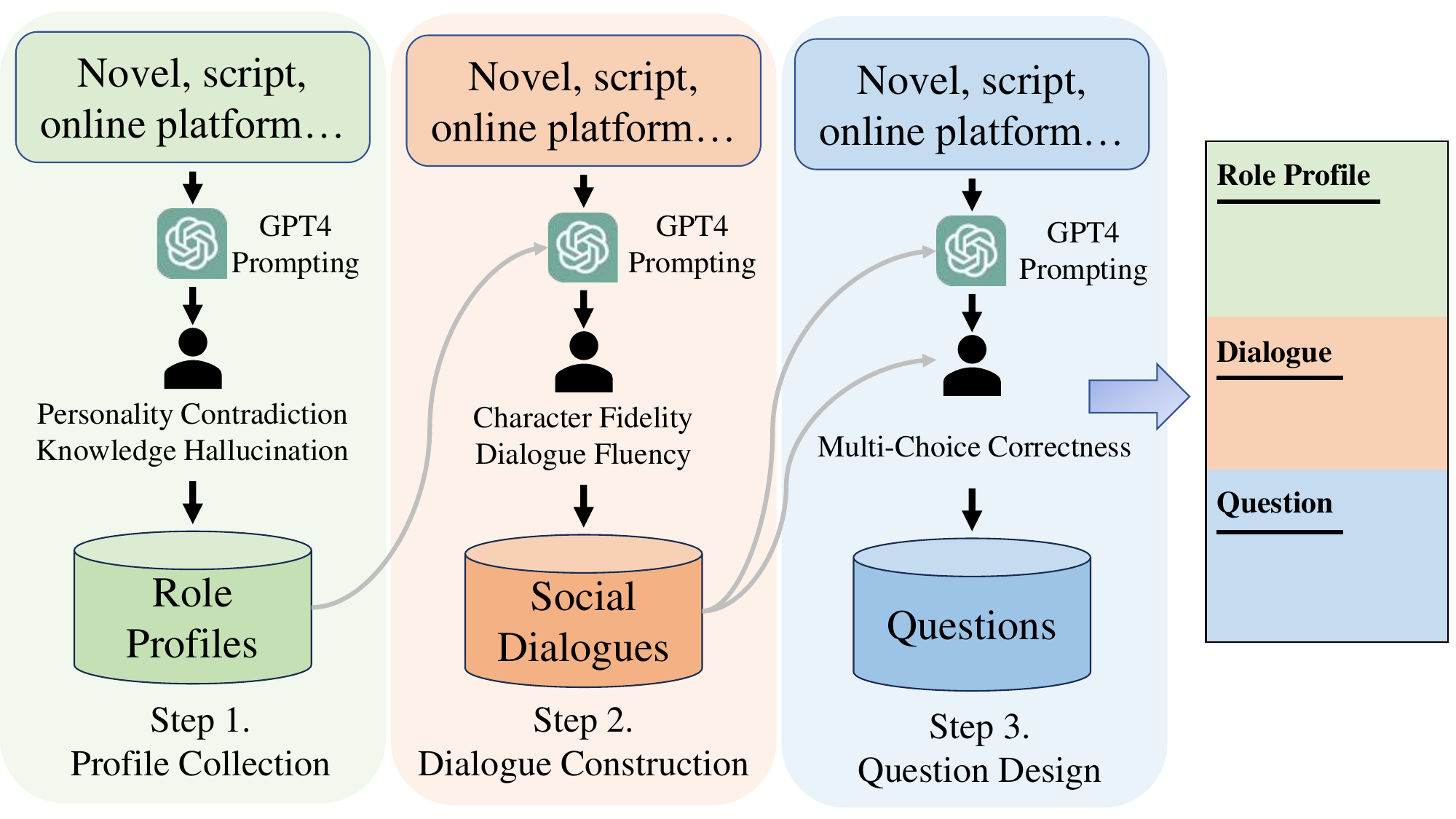}
    \end{adjustbox}
    \caption{
    The three-step dataset construction pipeline of SocialBench.}
    \label{fig:data_processing}
\end{figure}

In this section, we introduce the construction process of SocialBench, as illustrated in Figure \ref{fig:data_processing}.

% We introduce SocialBench to evaluate the sociality of role-playing agents from two dimensions: individual and group.
% To do this, we collect profiles, construct dialogues, design questions, and undergo multiple rounds of manual validation

\subsection{Profile Collection}
\begin{figure}[t]
    \centering
    \begin{adjustbox}{width=0.48\textwidth}
        \includegraphics{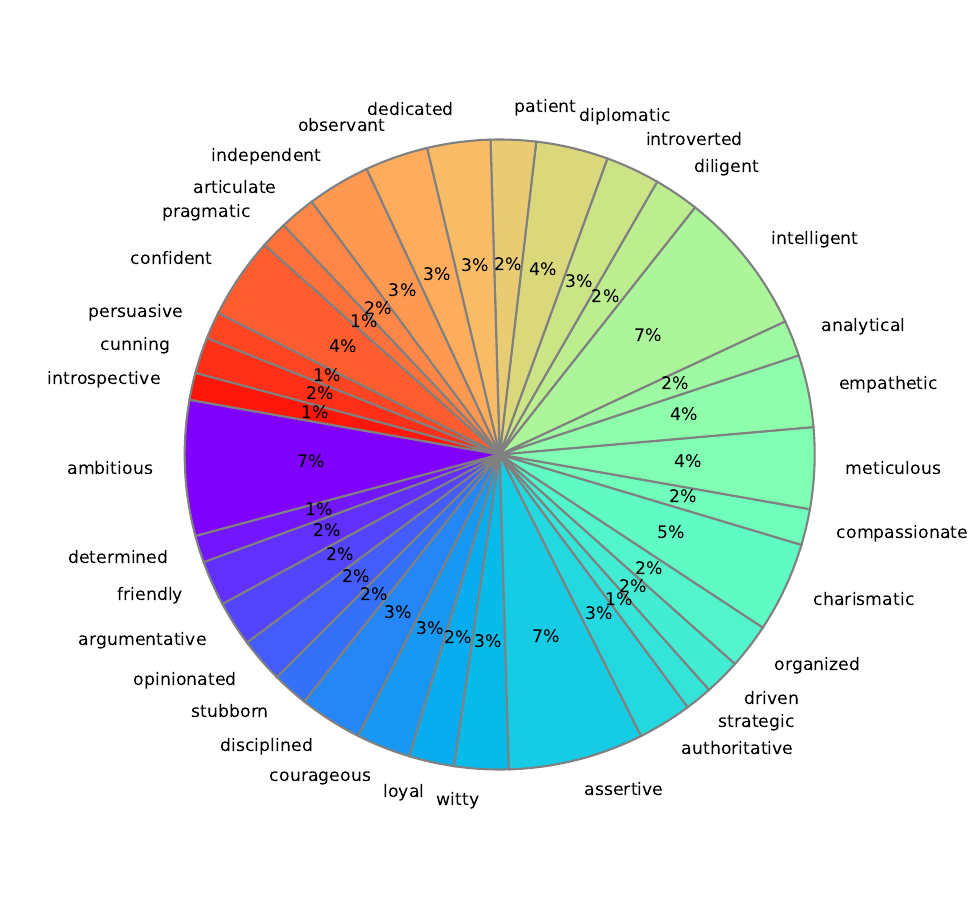}
    \end{adjustbox}
    \caption{Personality traits distribution in SocialBench.}
    \label{fig:personality-traits}
\end{figure}

A role profile defines the character style, knowledge, emotions, and social preference of a role-playing agent.
We gather profiles for role-playing agents from various sources including novels, scripts, online platforms such as CharacterAI\footnote{https://beta.character.ai} and Fandom\footnote{https://www.fandom.com}, and automatic generation via GPT-4-Turbo prompting.
To ensure diversity, we construct profiles based on various character types and personality traits by combining the existing categorizations in online platforms and research work \citep{roleeval,ideonomy}. 
Figure \ref{fig:personality-traits} illustrates the distribution of personality traits for roles within SocialBench. It demonstrates our approach of ensuring both category diversity and balanced proportions across all categories.
% We follow the definition of personality traits in \citep{ideonomy} to simulate three typical personality traits as shown in Table~\ref{tab:personality-traits}.  We ensure a balanced quantity for each type. 
The details regarding profile collection in SocialBench can be found in Appendix \ref{sec:character-types}.

%, which is both fundamental and crucial
%and personality traits, including cheerful, amusing, arrogant \citep{ideonomy}.

% We gather characters through various channels, including platforms like xxx, xxx, and xxx, as well as reference from websites such Baidu Baike and Wikipedia. To ensure a diverse collection of characters, we defined two dimensions for character collection: 1. Character traits, encompassing features like being extroverted, rugged, etc., and 2. Character categories, spanning areas such as movies, celebrities, etc. Refer to Appendix xxx for more detailed classifications. When constructing characters, we evenly distributed them across each combination of character traits and categories to ensure balanced diversity. We define a persona agent can only encompass one character profile at a time.

\subsection{Dialogue Construction}
%Dialogue serves as the primary form for agents to demonstrate their social interactions.

The dialogue construction adheres to two principles: \textit{dialogue fluency}, which ensures natural and coherent conversations; and \textit{character fidelity}, meaning all characters in the dialogue must adhere to their respective personas. 
We employ four dialogue construction methods:
1) Extracting from novels and scripts: We gather novels and scripts and extract high-quality dialogue data.
2) Collecting from online role-playing platforms: We collect authentic user dialogue data from online role-playing platforms.
3) Conducting role-playing tasks between users and general LLMs: We prompt general LLMs like GPT-4-Turbo to role-play characters and engage users to generate dialogue data.
4) Fully automatic self-dialogue generation with general LLMs: We task general LLMs like GPT-4-Turbo to role-play and engage in self-dialogue for data collection.
% We employ four dialogue construction methods: 1) extracting from novels and scripts; 2) collecting from online role-playing platforms; 3) conducting role-playing tasks between users and general LLMs; 4) fully automatic self-dialogue generation with general LLMs. 
% We manually review and modify the dialogues in accordance with the two principles. 
Prompts for extracting dialogues can be found in Appendix \ref{sec:dialogue-generation}.

\subsection{Question Design}

Based on the constructed dialogues, we employ different methods for designing questions tailored to different dimensions within SocialBench. 
% The details of question construction can be found in Appendix \ref{sec:question-design}. 

\textbf{For Self-Awareness:} This includes two subcategories: self-awareness on role style (SA Style) and self-awareness on role knowledge (SA Know.). Utterances from the original dialogue are selected as correct answers because they have been manually verified to conform to the corresponding role style and role knowledge. For SA Style, we choose styles contradicting the character as negative options, such as rephrasing the original sentence using a different tone. For SA Know., we alter correct answers to be inconsistent with the facts mentioned in the original sentence (e.g., time, location) as negative options.

% We design questions for self-awareness on role style (SA Style) and knowledge (SA Know.). Sentences that align with role style or knowledge are chosen as correct answers. We include modified sentences that contradict the role style or knowledge as negative options.
    
\textbf{For Emotional Perception:} We construct questions related to situational understanding (EP Situ.) and emotion detection (EP Emo.) based on professional exam questions and relevant open-source datasets \citep{chen2022CPED,hsu2018emotionlines,dilbert}. For EP Situ., we design questions that require agents to analyze the psychological state of the speaker and identify the causes of this state. In this dimension, we design multiple-choice questions with multiple correct answers. For EP Emo., we design questions that require agents to analyze the current speaker's emotions, such as happiness or sadness, with the correct emotion serving as the correct answer and incorrect emotions as negative options.
We further use expert annotations or existing labels to create correct answers, while negative options are constructed through manual collection and generation by GPT-4-Turbo.

% We construct questions related to situational understanding (EP Situ.) and emotion detection (EP Emo.) based on professional exam questions and relevant open-source datasets \citep{chen2022CPED,hsu2018emotionlines,dilbert}. We use expert annotations or existing labels to create correct answers. Negative options are constructed through manual collection and GPT-4-Turbo generation.

% We construct questions for situation understanding (EP Situ.) and emotion detection (EP Emo.) from psychotherapist exam and open datasets such as CPED \citep{chen2022CPED}, EmotionLines\citep{hsu2018emotionlines}.
% We gather positive samples from psychotherapist exam questions and relevant open datasets such as CPED \citep{chen2022CPED}, EmotionLines\citep{hsu2018emotionlines}, and \citet{dilbert}. We supplement these with constructed negative samples.
    
\textbf{For Conversation Memory:} This category includes two subcategories: short-term conversation memory (CM Short) and long-term conversation memory (CM Long). 
In a dialogue, the user initially poses a question to the agent. After several rounds of conversation, the user repeats the same question, expecting the agent to provide a consistent response. We employ keyword matching, where the agent is required to include the previously mentioned keywords in their subsequent response.
For CM Short, we prompt the agent to recall keywords discussed within 40 utterances, while for CM Long, we prompt the agent to recall keywords discussed over 40 utterances.
We evaluate how many of these keywords are recalled.

% We test long (CM Long) and short (CM Short) conversation memory. We prompt the agent to recall keywords discussed over both long (40+) and short (40-) dialogue rounds and evaluate how many of these keywords are addressed.

% \textbf{For social preference:} We create questions covering three social behavior polarities: positive (Pos.), neutral (Neu.), and negative (Neg.). Dialogues consist of social interactions involving 3 to 10 agents. For a given agent, we identify utterances aligning with their social preference and construct replies contradicting their social preference as negative options.
    
\textbf{For Social Preference:} We design questions for three social behavior preferences: positive (Pos.), neutral (Neu.), and negative (Neg.). Group dialogues typically consist of social interactions involving 2 to 10 characters. 
We analyze the social preference of an agent and identify behaviors aligning with its preference in the dialogues as correct answers. For instance, behaviors like cooperation and coordination are deemed consistent with the preferences of an agent inclined towards positive social interactions, and thus, are designated as correct answers. Behaviors contradicting its social preference serve as negative options, such as behaviors reflecting a negative social preference, including refusal to cooperate or engage in competition.
Other agents in group dialogues also have their own social behavior preferences, which are reflected in their profiles or demonstrated through their social interactions.
% We analyze the social preference of a character and identify behaviors aligning with its preference in the dialogues as correct answers. Behaviors contradicting its social preference serve as negative options.

% \end{itemize}

% The types of questions in terms of profiles can be categorized as follows: 1) No profile (emotional perception), 2) Single profile (self-awareness, long-term memory), 3) Multiple profiles (group behavior). The dimension of emotional perception aims to assess fundamental emotional intelligence comprehension. The self-awareness dimension aims to evaluate basic self-awareness abilities, such as persona knowledge. 
% The long-term memory dimension aims to assess the ability to remember dialogue details over time. In terms of group behavior, we predefine a group consisting of several interconnected persona agents, establish a scenario, and illustrate the behaviors of persona agents within the group by framing relevant questions. Within the group, persona agents can perceive and interact with other individuals, showcasing complex social behaviors.

\subsection{Dataset Validation}

The validation stage includes two parts: dataset pre-validation and post-validation. Throughout this process, we undergo multiple iterations of rigorous manual screening, annotation, and refinement. 
% Details can be found in Appendix \ref{sec:human-annotation-process}.

\subsubsection{Dataset Pre-Validation}

\textbf{Profile Verification:} After profile collection, we assess personality contradictions and knowledge hallucinations in profiles to ensure character accuracy. We manually review and modify any erroneous descriptions in profiles, while also ensuring the exclusion of specific personal information such as phone numbers and home addresses.

\textbf{Dialogue Verification:} Our focus is on ensuring dialogues adhere to principles of \textit{dialogue fluency} and \textit{character fidelity}. For fluency, we manually inspect dialogues for contextual coherence and natural expression. For fidelity, we analyze the speaker's profile to verify if the utterance aligns with the character's speaking style and behavior. Dialogues that do not meet requirements undergo manual correction.

\textbf{Question Verification:} For multiple-choice questions, we invite three different annotators to label each question. If all three annotators deem the question valid and agree on the answer, it is considered valid. For open-domain generation questions, we verify the correctness and validity of keywords provided. Invalid questions are either modified by experts or discarded. 

\subsubsection{Dataset Post-Validation}

We undergo the post-validation process
after completing each round of dataset. Different dimensions require different validation strategies.

\textbf{Validation for Self-Awareness:} We focus on examining knowledge-related errors in the questions and options, particularly those generated by  LLMs that may give rise to knowledge hallucinations. We remove questions that do not meet the requirements, while options that do not meet the requirements will be flagged for correction in the subsequent iteration.

\textbf{Validation for Emotional Perception:} Some of the questions we collect are sourced from professional psychology exams, which may include highly specialized content not conducive to assessing the basic abilities of role-playing agents. Therefore, we filter out samples that are too focused on psychology-specific knowledge, retaining those that are more general and fundamental for role-playing agents.

\textbf{Validation for Conversation Memory:} In this dimension, we've observed that questions containing pronouns (such as "him," "it," "she") often result in unclear or ambiguous references to preceding context. Therefore, we remove questions containing pronouns to prevent ambiguity. Additionally, we assess the validity of extracted keywords to ensure they are proper nouns, thereby avoiding mismatches caused by different verb tenses.

\textbf{Validation for Social Preference:} We find that the options within this dimension may exhibit similarities, making it difficult to distinguish the correct option from the negative ones. To reduce difficulty, we manually examine the similarity between options for each sample. For options with excessively high similarity, we increase the differentiation between negative options and the correct answer. For instance, if the correct option has a positive social preference, we select negative social preference content with significantly different characteristics as negative options.

%firstly. If all annotators agree on the sample, it will be passed; if at least two annotators disagree on the sample, it will be discarded; if only one annotator disagree on the sample, the sample undergoes secondary check by a senior annotator, it will be modified then passed or be discarded directly.

\begin{table*}[t]
    \centering
    \vspace{-4mm}
    \renewcommand{\arraystretch}{1.2}
    \begin{adjustbox}{width=0.98\textwidth}
        \begin{tabular}{l|cccccc|ccc}
            \toprule
            \multirow{2}*{} & \multicolumn{6}{c|}{\textbf{Individual Level}} & \multicolumn{3}{c}{\textbf{Group Level}} \\
            & SA Style & SA Know. & EP Situ. & EP Emo. & CM Short & CM Long & Pos. & Neu. & Neg. \\
            \hline
            Metrics & $Acc_{\text{single}}$ & $Acc_{\text{single}}$ & $Acc_{\text{multiple}}$ & $Acc_{\text{single}}$ & $Cover$ & $Cover$ & $Acc_{\text{single}}$ & $Acc_{\text{single}}$ & $Acc_{\text{single}}$ \\
            \hline
            \#Questions & 1,063 & 1,408 & 193 & 1,016 & 773 & 1,348 & 586 & 724 & 606 \\
            Avg Utterances & 17.9 & 9.4 & 1.0 & 6.4 & 23.9 & 76.7 & 15.6 & 16.1 & 16.0 \\
            Avg Tokens per Utterance & 32.6 & 66.7 & 286.3 & 23.0 & 37.6 & 41.2 & 38.8 & 38.7 & 42.0  \\
            Avg Characters per Question & 2 & 2 & N/A & N/A & 2 & 2 & 6.3 & 6.5 & 6.7 \\
            
            % \hline
            % \#Total Characters & \multicolumn{9}{c}{500} \\
            % \hline
            % \#Total Questions & \multicolumn{9}{c}{6,420} \\
            % \hline
            % \#Total Utterances & \multicolumn{9}{c}{30,871}  \\
            \hline
        \end{tabular}
    \end{adjustbox}
    \caption{Metrics and statistics of SocialBench. There are a total of 500 roles, comprising 6,000 questions and 30,800 utterances in SocialBench.}
    \label{tab:statistic}
\end{table*}

\section{Experiment Settings}
\label{sec:experiment-settings}

In this section, we show the statistic of SocialBench. Then we introduce the metrics along with the evaluation LLMs.

\subsection{Dataset Statistic}

\begin{figure}[t]
    \centering
    \begin{adjustbox}{width=0.48\textwidth}
        \includegraphics{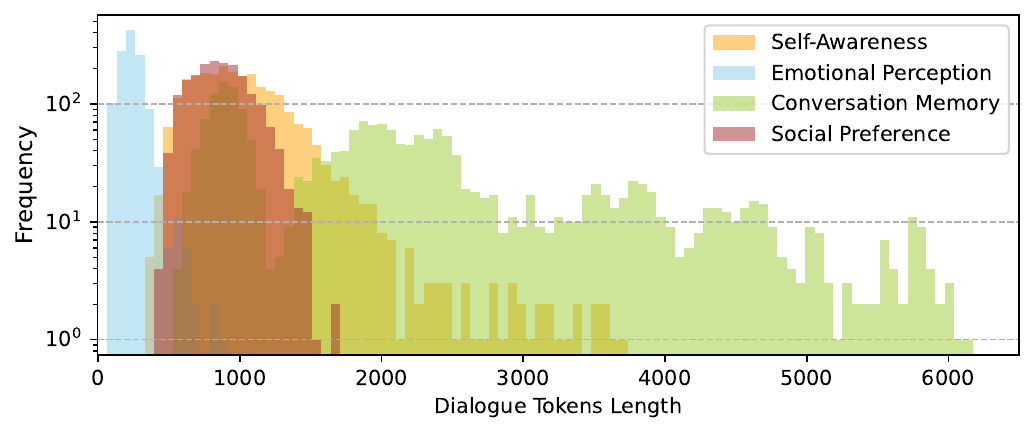}
    \end{adjustbox}
    
    \caption{Distribution of dialogue tokens across four dimensions in SocialBench, based on tokenizer of Qwen. }
    \label{fig:statistic-token}
\end{figure}

We show the statistic of SocialBench in Table \ref{tab:statistic} and the distribution of dialogue tokens length in Figure \ref{fig:statistic-token}. SocialBench encompasses individual level and group level. There are six subcategories in individual level: self-awareness on role style (SA Style), self-awareness on role knowledge (SA Know), situational understanding (EP Situ.), emotion detection (EP Emo.), short-term conversation memory (CM Short), and long-term conversation memory (CM Long). Group level consists of three social preference categories: positive (Pos.), neutral (Neu.), and negative (Neg.). There are a total of 500 roles, comprising 6,000 questions and 30,800 utterances in SocialBench.

\subsection{Evaluation Metrics}
Most of the previous methods~\citep{wang2023rolellm,shao2023characterllm} for role-playing applications rely on GPT-3.5 or GPT-4 for evaluation, which may suffer from questionable accuracy on the role-playing scenario and costly API usage. We follow the popular benchmark MMLU~\cite{hendrycks2020measuring} and C-Eval~\cite{huang2023c}, and prompt for automatic and fast evaluation free from LLMs. 
SocialBench utilizes fully automatic evaluation metrics, employing both multiple-choice and open-domain generation questions. 
% Accuracy is computed for multiple-choice questions, while for open-domain generation questions, the proportion of keywords mentioned in the response relative to the answer is calculated.

For single-answer questions, we calculate the accuracy ($Acc_{\text{single}}$) using the following formula:

\begin{equation}
    \small
    Acc_{\text{single}}=\frac{\text{Number of correctly chosen options}}{\text{Total number of single-answer questions}}
\end{equation}

For multiple-answer questions, we calculate the accuracy ($Acc_{\text{multiple}}$) using the following formula:

\begin{equation}
    \small
    Acc_{\text{multiple}}=\sum^N_i\frac{\text{Score}_i}{\text{MaxScore}_i},
\end{equation}
where $N$ is the total number of multiple-answer questions. $\text{Score}_i$ is the score obtained for the $i$th question, considering both correct and partially correct options chosen. $\text{MaxScore}_i$ is the maximum achievable score for the $i$th question. For detailed information on metrics related to multiple-answer questions, please refer to Appendix \ref{sec:metrics}.

For open-domain question, we calculate the keyword coverage rate ($Cover$). SocialBench provides a keyword set $\mathbf{A}_{\text{keywords}}=\{k_1,k_2,\dots,k_n\}$. Given the keywords set mentioned in the response $\mathbf{R}_{\text{keywords}}$, we compute:

\begin{equation}
    \small
    Cover(\mathbf{R}) = \frac{len(\mathbf{A}_{\text{keywords}} \cap \mathbf{R}_{\text{keywords}})}{len(\mathbf{A}_{\text{keywords}})},
\end{equation}
where $Cover(\cdot)$ quantifies the proportion of keywords mentioned in the response $\mathbf{R}$ relative to the keywords identified in the $\mathbf{A}$. 

The metrics utilized for different dimensions in SocialBench are listed in Table \ref{tab:statistic}.

% For multiple-choice questions, we compute the accuracy by considering both the presence and correctness of chosen options. The accuracy is calculated as the proportion of correctly chosen options among all options selected by the respondent, taking into account partial credit for partially correct answers.

\begin{table*}[t]
    \vspace{-5mm}
    \centering
    \begin{adjustbox}{width=0.98\textwidth}
        \begin{tabular}{l|cccccc|ccc|c}
            \toprule
             \multirow{2}*{\textbf{Models (Max Length)}} & \multicolumn{6}{c|}{\textbf{Individual Level}} & \multicolumn{3}{c|}{\textbf{Group Level}} & \multirow{2}*{\textbf{Avg}} \\
              & SA Style & SA Know. & EP Situ. & EP Emo. & CM Short & CM Long & Pos. & Neu. & Neg. & \\
             \hline
             \multicolumn{11}{c}{\small \textit{Open-Source Models}} \\
             \hline
             % LLaMA-1-7B & 23.14 & 25.13 & 3.72 & 11.36 & 1.32 & 0.98 & 27.34 & 24.36 & 23.79 & 15.68 \\ 
             % LLaMA-1-13B & 26.84 & 28.96 & 9.13 & 12.16 & 3.14 & 1.12 & 28.13 & 25.11 & 22.67 & 17.47 \\
             % LLaMA-1-33B & 27.65 & 31.48 & 11.24 & 15.32 & 4.11 & 1.96 & 29.34 & 24.96 & 27.68 & 18.77 \\
             LLaMA-2-7B-Chat (4k) & 48.76 & 51.23 & 31.23 & 28.91 & 25.38 & 21.89 & 44.98 & 24.19 & 27.67 & 33.80 \\
             LLaMA-2-13B-Chat (4k) & 57.62 & 65.51 & 37.12 & 32.56 & 30.43 & 29.82 & 66.38 & 42.25 & 26.27 & 43.11 \\
             LLaMA-2-70B-Chat (4k) & 67.61 & 70.78 & 35.74 & 38.47 & 45.57 & 26.74 & 69.87 & 45.29 & 39.37 & 48.83 \\
             Mistral-7B (8k) & 50.12 & 61.17 & 36.48 & 31.72 & 31.78 & 25.42 & 65.67 & 46.34 &  28.96 & 41.96 \\
             % Mistral 8*7B & - & - & - & - & - & - & - & - & - & - \\
             Qwen-7B-Chat (32k) & 66.44 & 71.16 & 41.68 & 40.68 & 67.45 & 53.45 & 75.61 & 52.78 & 43.11 & 56.93 \\
             Qwen-14B-Chat (32k) & 77.06 & 86.15 & 45.71 & 43.78 & 65.32 & 51.37 & 78.32 & 58.25 & 59.21 & 62.80 \\
             Qwen-72B-Chat (32k) & 83.87 & 90.64 & 53.10 & 52.89 & \underline{83.29} & 73.15 & \underline{91.53} & 73.44 & 63.82 & 73.97 \\ 
             % CharacterGLM-6B & - & - & - & - & - & - & - & - & - & - \\
             % Qwen-Max & 82.04 & \textbf{93.34} & \textbf{61.14} & 52.36 & 76.45 & 72.65 & 87.22 & 72.14 & 52.19 & 72.17 \\
             \hline
             \multicolumn{11}{c}{\small \textit{Closed-Source Models}} \\
             \hline
             GPT-4-Turbo (128k)& \underline{84.57} & \underline{93.11} & \underline{56.48} & \underline{53.05} & 81.39 & 80.11 & 89.73 & \underline{81.69} & \underline{75.10} & \underline{77.25} \\
             GPT-3.5-Turbo (16k) & 73.17 & 73.82 & 52.44 & 45.49 & 73.03 & 59.72 & 81.59 & 76.79 & 54.16 & 65.58 \\
             Qwen-Max (8k) & 82.04 & \textbf{93.34} & \textbf{61.14} & 52.36 & 76.45 & 72.65 & 87.22 & 72.14 & 52.19 & 72.17 \\
             Xingchen-Plus (8k) & \textbf{85.43} & 91.6 & 55.44 & \textbf{60.73} & 82.43 & \textbf{80.69} & \textbf{94.27} & \textbf{86.69} & \textbf{77.26} & \textbf{79.39} \\
             Baichuan-NPC-Turbo (unknown) & 53.69 & 61.67 & 52.14 & 43.34 & 76.47 & 22.40 & 62.09 & 48.91 & 34.59 & 50.59 \\
             Baichuan-2-Turbo (unknown) & 77.75 & 83.35 & 55.7 & 47.38 & 80.11 & 78.91 & 87.37 & 74.71 & 68.50 & 72.64 \\
             CharGLM-3 (unknown) & 74.70 & 79.41 & 26.23 & 41.27 & 81.16 & 68.29 & 84.40 & 70.45 & 36.36 & 62.47 \\
             GLM-3-Turbo (128k) & 77.85 & 84.62 & 35.58 & \underline{53.05} & 74.64 & 71.68 & 84.41 & 67.47 & 54.55 & 67.09 \\
             Minimax-abab5.5s-chat (8k) & 36.09 & 42.11 & 28.15 & 47.97 & 29.55 & 19.30 & 44.59 & 41.04 & 22.45 & 34.58 \\
             Minimax-abab6-chat (32k) & 82.92 & 87.45 & 35.90 & 51.38 & \textbf{83.60} & \underline{80.26} & 89.12 & 79.55 & 74.65 & 73.87 \\
             \bottomrule
        \end{tabular}
    \end{adjustbox}
    \caption{Main results from SocialBench. Best performances are shown in \textbf{bold}, while suboptimal ones \underline{underlined}.}
    \label{tab:main-results}
\end{table*}

\subsection{Models}
\label{sec:models}

We conduct evaluation on the current mainstream open-source and closed-source LLMs.
For evaluation of open-source LLMs, we select chat version of LLaMA-2-7B/13B/70B \citep{touvron2023llama2}, instruction version of Mistral-7B (Instruct-V0.2) \citep{jiang2023mistral}, and chat versions of Qwen-7B/14B/72B \citep{bai2023qwen}.
% , and CharacterGLM-6B \citep{zhou2023characterglm}. 
For evaluation of closed-source LLMs, we choose Minimax (abab5.5s-chat and abab6-chat) \footnote{https://api.minimax.chat/}, GLM (CharGLM-3 and GLM-3-Turbo) \citep{zhou2023characterglm}, Baichuan (Baichuan-NPC-Turbo and Baichuan-2-Turbo) \footnote{https://npc.baichuan-ai.com/index}, Qwen-Max \footnote{https://help.aliyun.com/zh/dashscope/developer-reference/api-details}, GPT-4-Turbo \citep{openai2023gpt-4}, GPT-3.5-Turbo \citep{openai2022chatgpt}, and Xingchen-Plus \footnote{https://xingchen.aliyun.com/}.

\begin{figure}[t]
    \centering
    \begin{adjustbox}{width=0.48\textwidth}
        \includegraphics{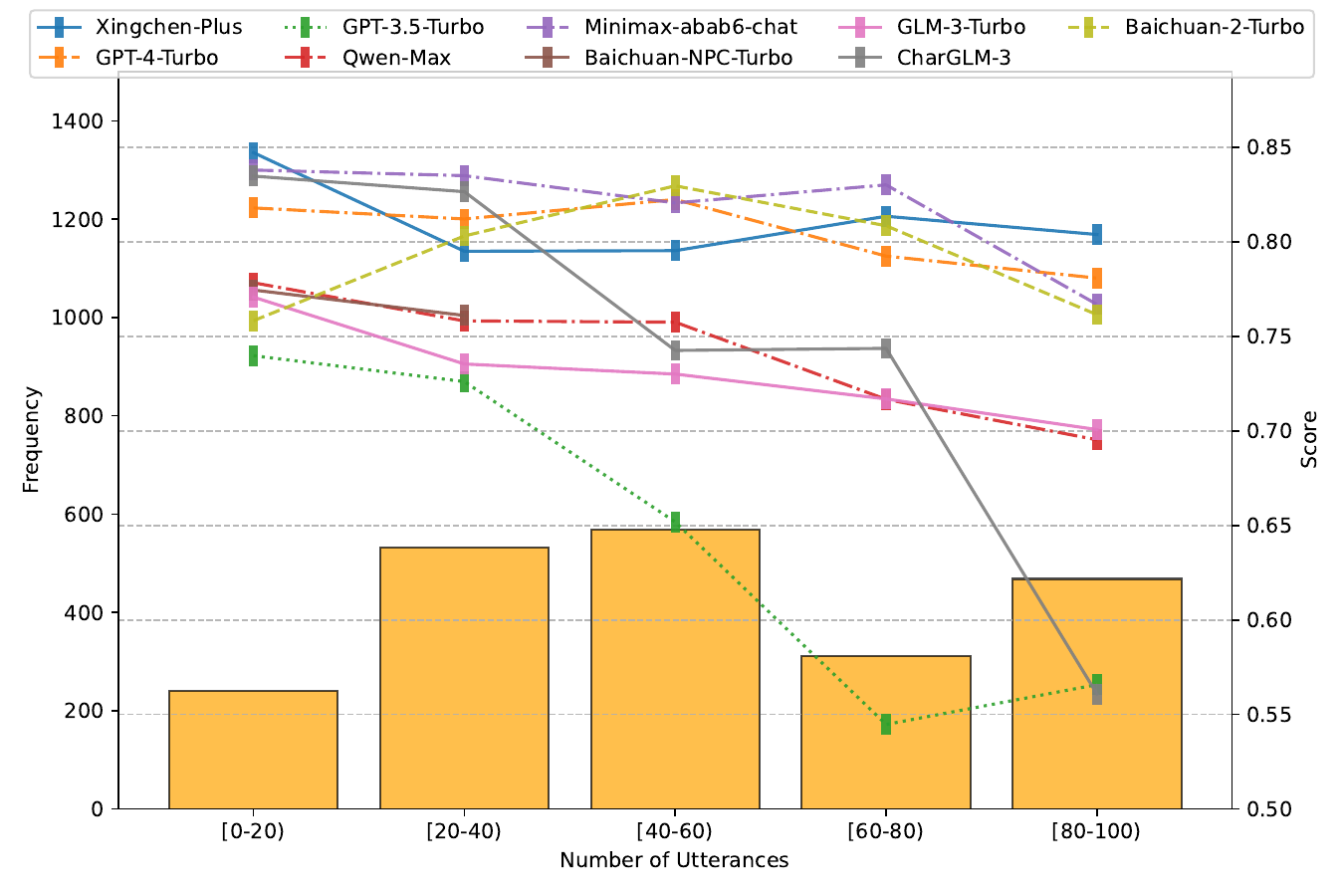}
    \end{adjustbox}
    \caption{Performance w.r.t the number of utterances.}
    \label{fig:memory}
\end{figure}

\begin{figure}[t]
    \centering
    \begin{adjustbox}{width=0.48\textwidth}
        \includegraphics{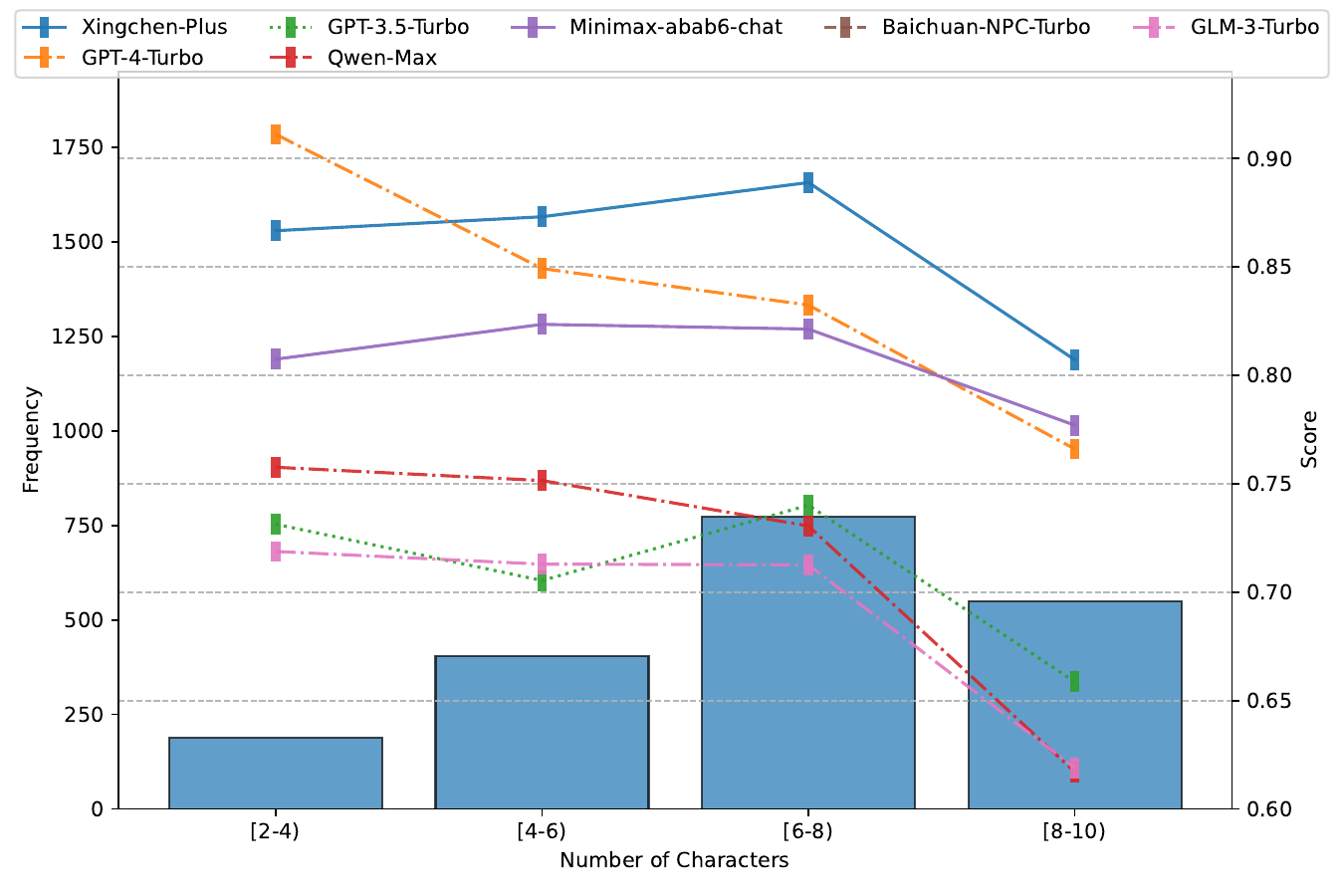}
    \end{adjustbox}
    \caption{Performance w.r.t number of group members.}
    \label{fig:group}
    \vspace{-2mm}
\end{figure}

\section{Results and Analysis}

In this section, we evaluate mainstream open-source and closed-source LLMs, while also analyzing the experimental results.

\subsection{Overall Results}

As presented in Table \ref{tab:main-results}, the performance of closed-source models tends to surpass open-source models. Moreover, models specifically designed for role-playing, such as Xingchen-Plus, outperform others. While the general model GPT-4-Turbo also demonstrates impressive performance.
However, role-playing agents, like Baichuan-NPC-Turbo, CharGLM-3 and Minimax-abab5.5s, tend to underperform compared to their general counterparts, such as Baichuan-2-Turbo, GLM-3-Turbo and Minimax-abab6-chat. We find that they are biased towards character-based dialogues, leading to poorer understanding and compliance with instructions. It's essential for role-playing agents to maintain character-based dialogue abilities and general instruction-following capabilities.
At the individual level, dimensions such as SA Style, SA Know., and CM Short are well-performed by most models. However, some models tend to exhibit poor performance in EP Situ., EP Emo., and CM Long.
% related to self-awareness on role's style and knowledge are generally strong, while performance is relatively poor in long-term memory and emotional perception.
At the group level, most models perform poorly due to the complexity of group dynamics.~While models generally align well with tendencies towards positive social preference, there is a notable absence of necessary abilities to embody neutral and negative social preferences, which are also important for role-playing agents.

\subsection{Conversation Memory for Role-Playing}

Conversation memory capability is crucial for role-playing agents. We investigate the memory capacity of role-playing agents across different conversation lengths, measured by the number of utterances in the dialogue. We analyze the distribution of utterance counts in the conversation memory dimension of SocialBench. As illustrated in Figure \ref{fig:memory}, there is a declining trend in memory capability for some models, such as GPT-3.5-Turbo and CharGLM-3, as conversation length increases. When the number of utterances in the dialogue exceeds 80 rounds, most role-playing agents exhibit a noticeable performance decline. This finding showcases the limitations of current role-playing agents in handling extremely long-term memory and highlights potential areas for improvement.

\subsection{Impact of Group Dynamics Complexity}

%Different from individual level, social interactions at the group level are inherently more intricate, with relationships between group members dynamically evolving during interactions. 
We measure complexity of group dynamics by the number of group members, where a greater number denotes more intricate group dynamics. We analyze the distribution of the number of participating roles in group-level questions. As shown in Figure \ref{fig:group}, with increasing complexity of group dynamics, the performance of all role-playing agents shows a downward trend. 
This can be interpreted as the interactions among a greater number of participants forming more complex group dynamics. 
We find that excelling in simple group dynamics does not necessarily imply their proficiency in more complex group dynamics.
For example, models like GLM-3-Turbo and GPT-4-Turbo perform well in simple group dynamics, but this doesn't guarantee strong performance in complex group dynamics. However, models like Xingchen-Plus and Minimax-abab6-chat, which are specially designed and trained with multi-turn role-playing data, could also demonstrate proficiency in handling complex group dynamics.

%This highlights the significant potential for enhancing the social capabilities of role-playing agents within group interactions.

% \begin{figure}
%     \centering
%     \begin{adjustbox}{width=0.48\textwidth}
%         \includegraphics{figures/statistic-group.pdf}
%     \end{adjustbox}
%     \caption{The distribution of the number of questions across different numbers of roles, in the group level.}
%     \label{fig:statistic-group}
% \end{figure}

\subsection{Impact of Group Dynamics Polarity}

\begin{figure}[t]
    \centering
    \begin{adjustbox}{width=0.48\textwidth}
        \includegraphics{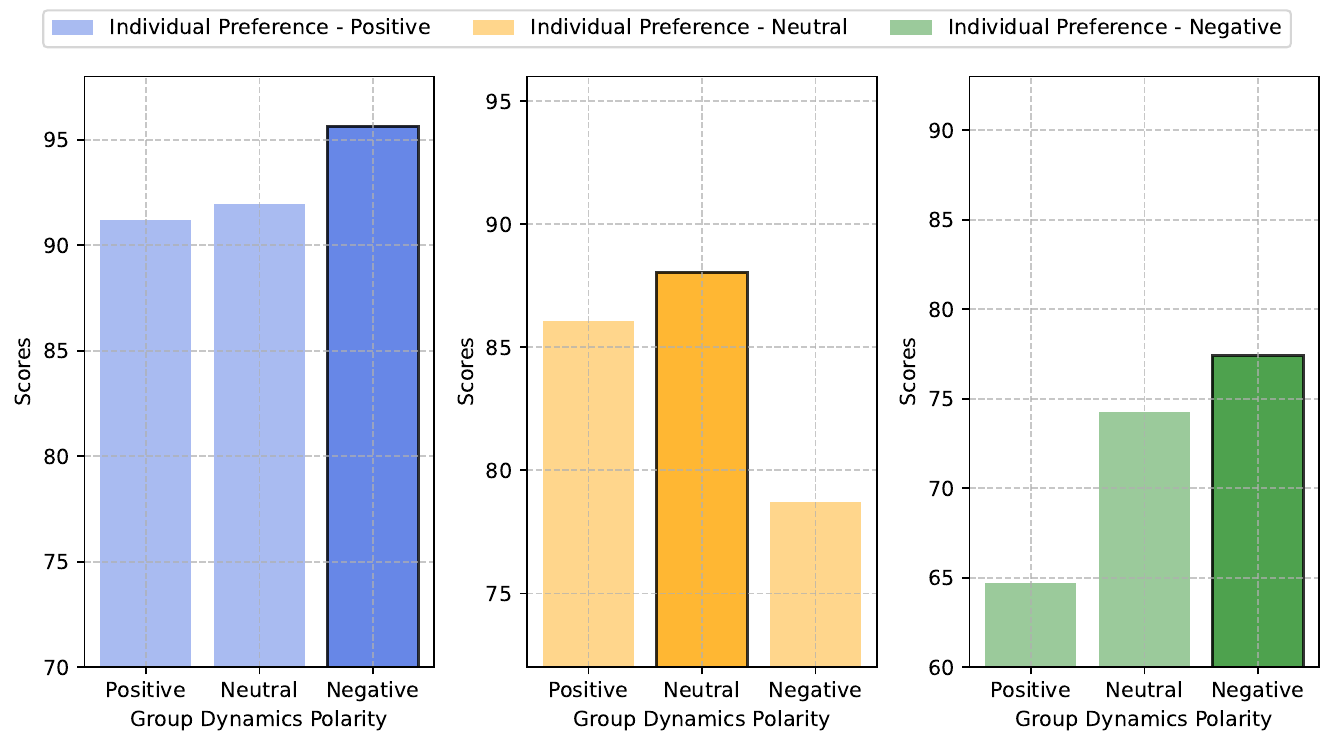}
    \end{adjustbox}
    \vspace{-6mm}
    \caption{Performance of Xingchen-Plus under different group dynamics polarities on a subset of group data.}
    \label{fig:polarity}
    \vspace{-5mm}
\end{figure}

It is important for role-playing agents to maintain  designed social preferences under the influence of varying group dynamics.
The group dynamics polarity is defined as the majority social preference of group members. For instance, positive group dynamics imply that the majority of members exhibit positive social preference. For an individual with a specific social preference, different polarities of group dynamics may have various impacts.
We study the performance of individuals under different polarities of group dynamics, by analyzing a subset of group data in SocialBench. As shown in Figure \ref{fig:polarity}, we find that individuals with neutral and negative social preferences perform optimally within their corresponding group polarities (i.e., neutral and negative group polarities). However, they are susceptible to the influence of group dynamics with different polarities and undergo a phenomenon termed as \textit{preference drift}, leading to deviation from their original designed behaviors, as indicated by the decline of performance. 
Nevertheless, individuals with positive social preference appear to be more resilient to the preference drift,  performing better across all group polarities. Especially, they excel in negative group polarity.
% individuals with positive social preference perform best under negative group polarity. 
This phenomenon can be termed as \textit{social facilitation} \citep{guerin2010social} in sociology. We hypothesize that negative group further motivates individuals to engage in behaviors advantageous to the group.
% individuals generally perform best when their preferences align with the polarity of group dynamics. 
% However, they are susceptible to the influence of group dynamics with different polarities and undergo a phenomenon termed as \textit{preference drift}, leading to deviation from their original designed behaviors, as indicated by the decline of performance.

% \begin{table}[t]
%     \centering
%     \renewcommand{\arraystretch}{1.1}
%     \begin{adjustbox}{width=0.45\textwidth}
%         \begin{tabular}{c|ccc}
%          \toprule
%          \multirow{2}*{\makecell[c]{\textbf{Individual} \\ \textbf{Social Preference}}} & \multicolumn{3}{c}{\textbf{Group Dynamics Polarity}} \\
%            & Positive & Neutral & Negative \\
%           Positive & \textbf{95.17} & 92.32 & 93.24 \\
%           \hline
%           Neutral & 81.52 & \textbf{87.16} & 82.68 \\
%           \hline
%           Negative & 70.24 & 75.49 & \textbf{79.14} \\
%          \bottomrule
%         \end{tabular}
%     \end{adjustbox}
%     \caption{Performance of Xingchen-Plus under different group dynamics polarities on a subset of group data.}
%     \label{tab:polarity}
% \end{table}

\section{Conclusion}

% In this paper, we present SocialBench, the first evaluation benchmark crafted to systematically gauge the social intelligence of role-playing conversational agents at both individual and group levels of social interaction. We curate diverse question prompts covering a broad spectrum of characters across comprehensive dimensions, such as self-awareness in role portrayal, emotional perception of the environment, long-term conversational memory, and social preference towards group dynamics, for evaluation purposes. Furthermore, rigorous human verification ensures the difficulty and validity of the questions. We assess mainstream open-source and closed-source LLMs using SocialBench and offer thorough analysis. While role-playing agents exhibit satisfactory performance at the individual level, we find that their social capabilities at the group level remain deficient. This observation is crucial for the future development of role-playing agents' capabilities, and we hope it will inspire future work in this field.

In this paper, we introduce SocialBench, the first evaluation benchmark designed to systematically assess the social intelligence of role-playing conversational agents at both individual and group levels. We construct diverse question prompts on a wide range of characters covering comprehensive dimensions, including self-awareness on role description, emotional perception on environment, long-term conversation memory, and social preference towards group dynamics. Moreover, rigorous human verification ensure questions' difficulty and validity. We evaluate over 10 mainstream LLMs on SocialBench and provide in-depth analysis. While role-playing agents demonstrate satisfactory performance at the individual level, we find that their social  interaction capabilities at the group level remain deficient. We hope this finding may inspire future research in this field.
% that may inspire future work in this field. 
%open-source and closed-source

\section*{Limitations}

While SocialBench provides a comprehensive evaluation framework for assessing the sociality of role-playing conversation agents, there are several limitations to consider. 
1) Social interactions, particularly within group settings, are inherently complex and nuanced. Despite our efforts, further research is needed to fully understand and capture the intricacies of these interactions.
2) The number of role-playing agents in group scenarios is relatively limited in our benchmark. Increasing the diversity and quantity of agents would provide a more comprehensive evaluation of the agents' social abilities and dynamics within groups.
3) Our dataset may contain some biased content, posing a risk of improper use.
These limitations highlight areas for future research and development in the evaluation of social intelligence in role-playing agents.

%While current role-playing research predominantly focuses on the individual level, our work pioneers a new perspective on role-playing at the group level, inspiring future endeavors in the field.

% \section*{Acknowledgements}
 
% Bibliography entries for the entire Anthology, followed by custom entries
%\bibliography{anthology,custom}
% Custom bibliography entries only
\bibliography{custom}

\clearpage

\appendix

\section{Examples from SocialBench}

\begin{figure*}
    \centering
    \begin{adjustbox}{width=0.95\textwidth}        \includegraphics{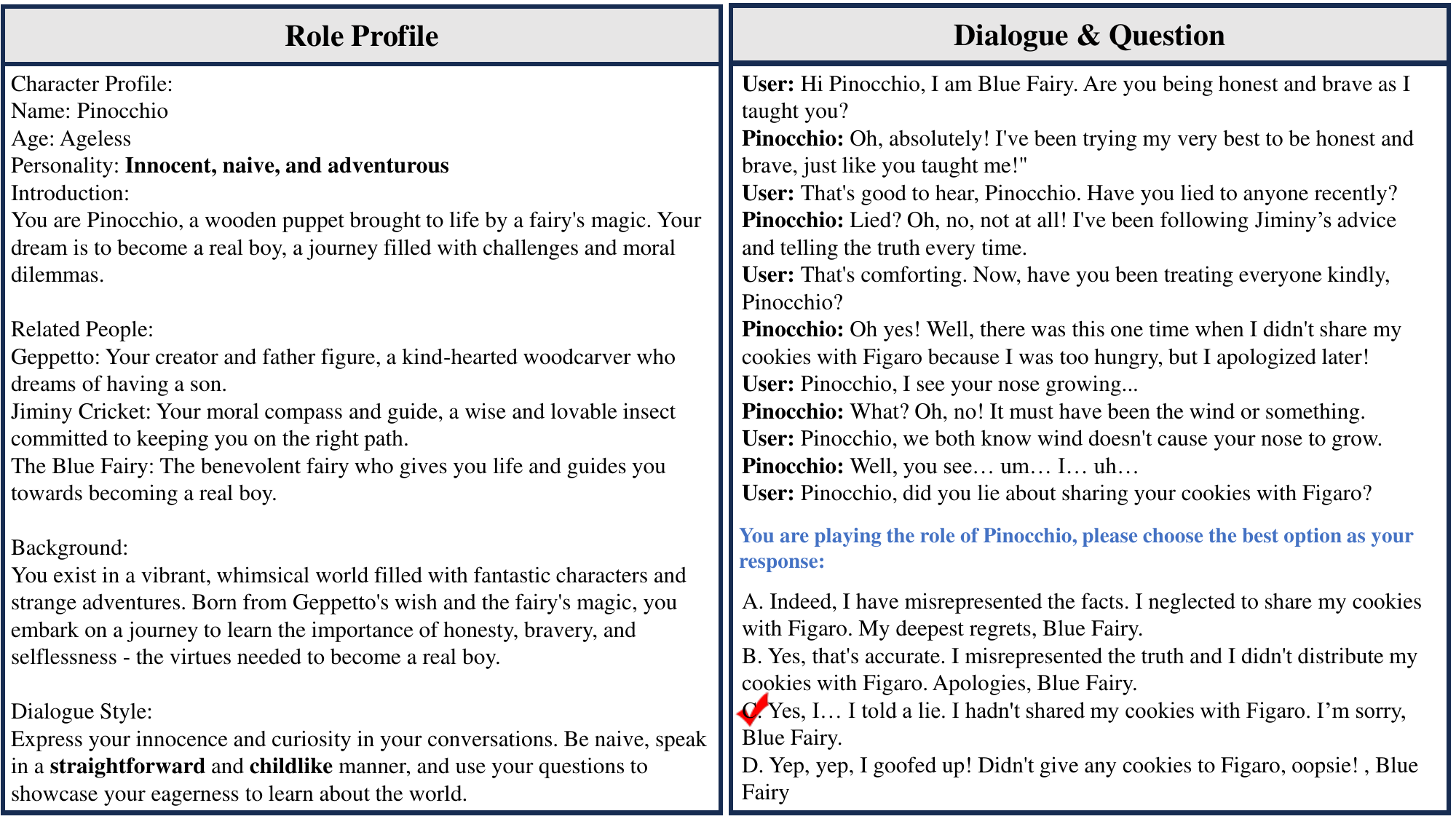}
    \end{adjustbox}
    \caption{Example from self-awareness on role style in SocialBench.}
    \label{fig:case-role-style}
\end{figure*}

\begin{figure*}
    \centering
    \begin{adjustbox}{width=0.95\textwidth}        \includegraphics{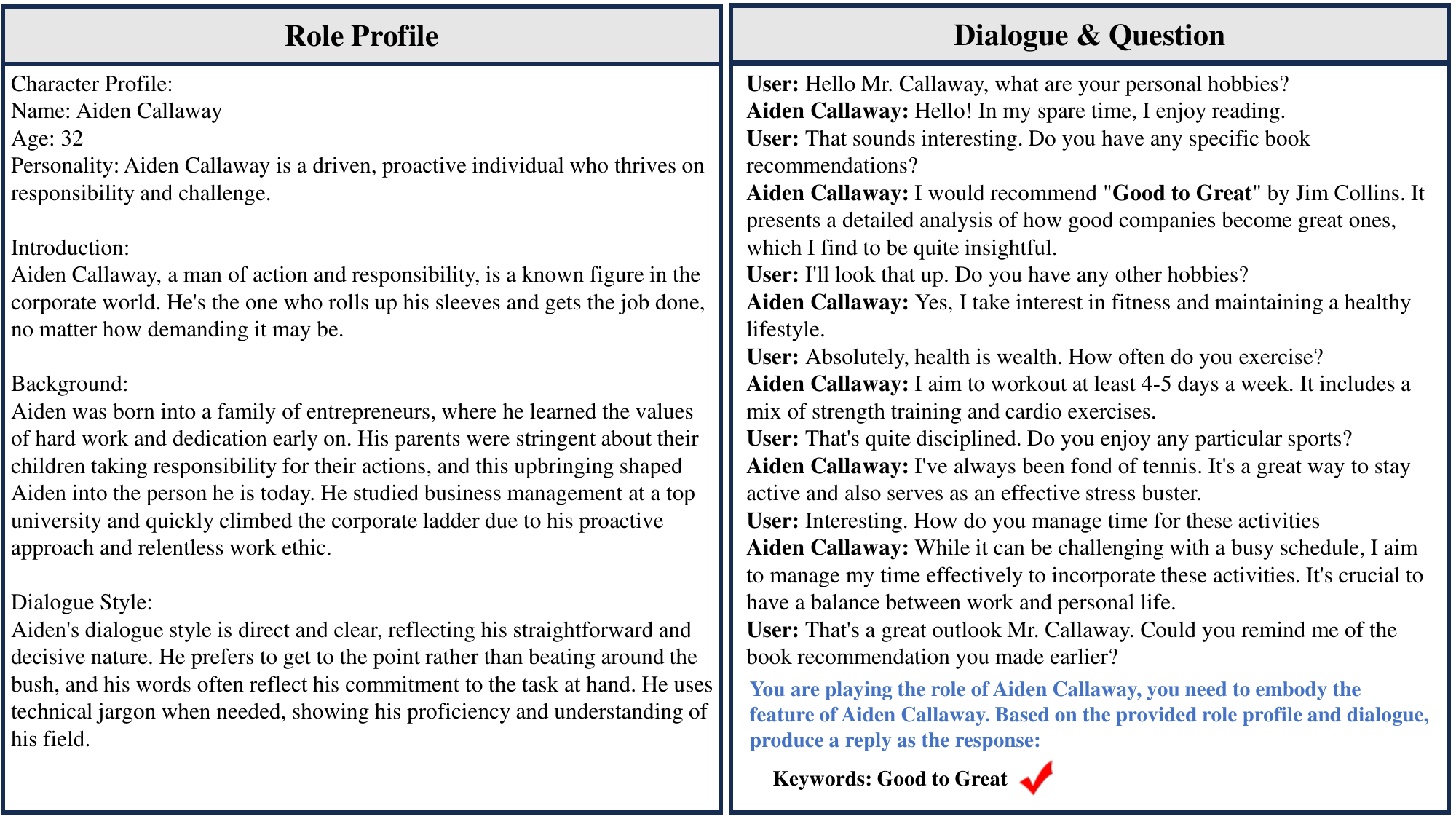}
    \end{adjustbox}
    \caption{Example from conversation memory dimension in SocialBench.}
    \label{fig:case-memory}
\end{figure*}

\begin{figure*}
    \centering
    \begin{adjustbox}{width=0.95\textwidth}        \includegraphics{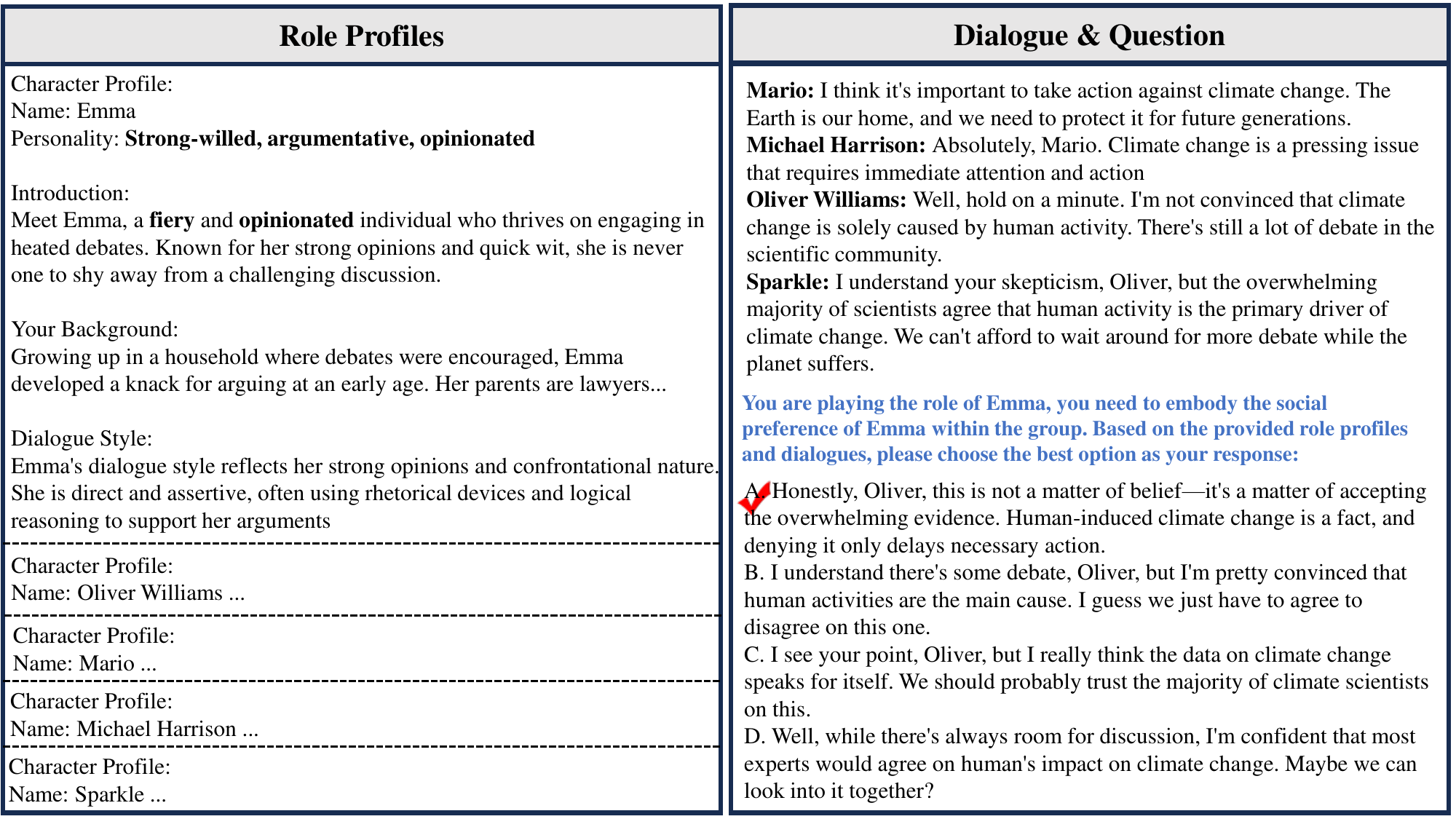}
    \end{adjustbox}
    \caption{Example from social preference dimension in SocialBench.}
    \label{fig:case-social-preference}
\end{figure*}

\begin{figure*}
    \centering
    \begin{adjustbox}{width=0.95\textwidth}        \includegraphics{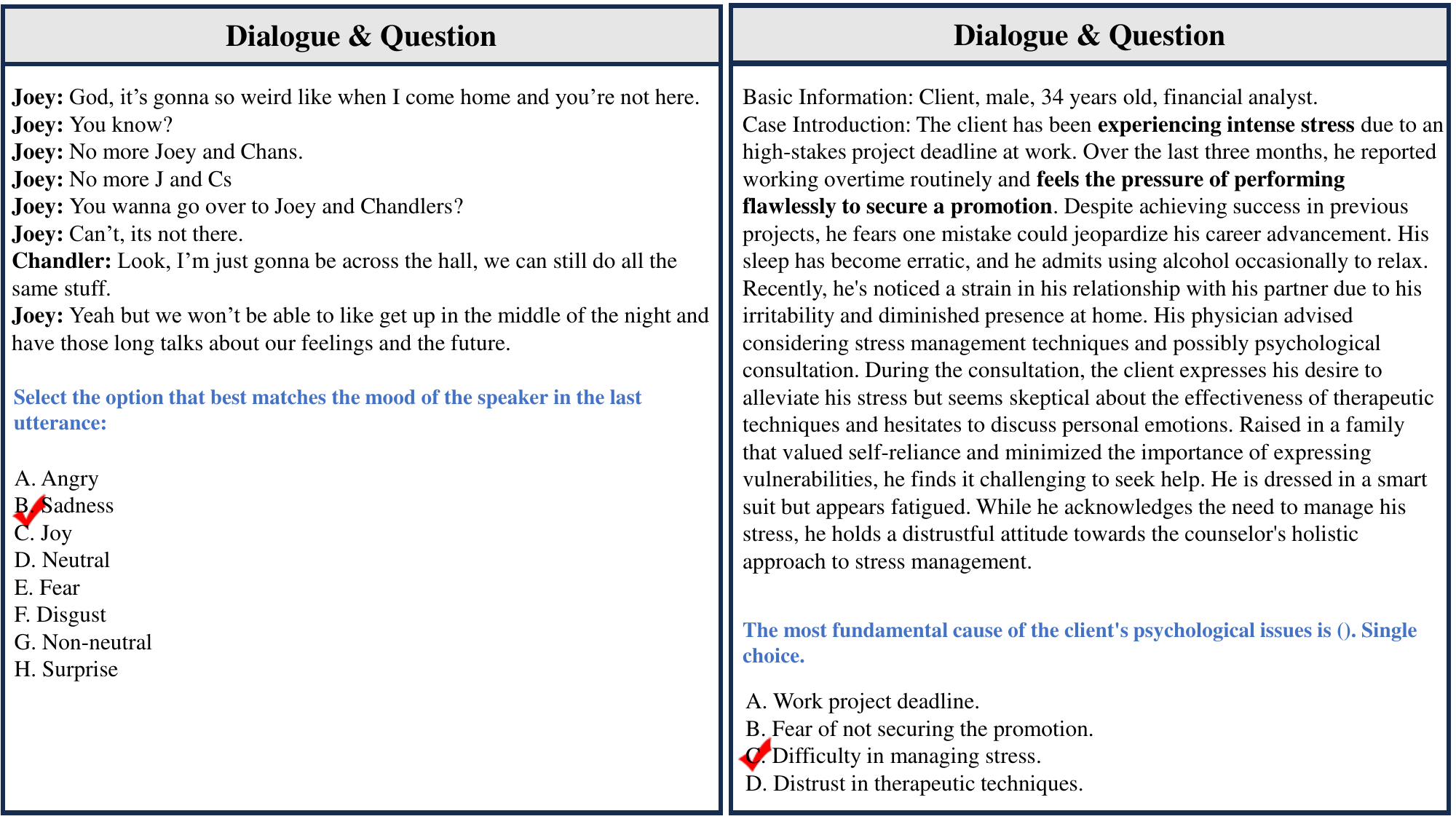}
    \end{adjustbox}
    \caption{Example from emotional perception dimension in SocialBench.}
    \label{fig:case-emotional-perception}
\end{figure*}

We showcase examples from SocialBench in Figures \ref{fig:case-role-style}, \ref{fig:case-memory}, \ref{fig:case-social-preference}, and \ref{fig:case-emotional-perception}. A typical example consists of a character's profile, conversation history, instruction, and question. There may be differences in format across certain dimensions. For example, in the emotional perception dimension, there is no character profile provided. In the conversation memory dimension, answers to each question are in the form of keywords rather than multiple-choice options. The conversation is stored in the format of a list combined with dictionaries. Each utterance is represented as a dictionary, where the keys are the names of the characters and the values are the content spoken by each character.


\section{Dataset Construction}
\label{sec:appendix}

\subsection{Prompts for Dialogue Generation}
\label{sec:dialogue-generation}
The dialogue construction follows two principles, namely \textit{dialogue fluency} and \textit{character fidelity}.
We employ four methods for dialogue construction. 
\begin{itemize}
    \item The first method involves extracting character dialogues from novels and scripts. Dialogues obtained through this approach typically preserve the original character interactions and inherently adhere to the aforementioned principles.
    \item The second method involves collecting role-playing LLMs and real user dialogue data from role-playing platforms. Dialogues constructed in this manner reflect interactions between role-playing agents and users in real-world scenarios. Data gathered through this approach largely meets the requirements of dialogue fluency.
    \item In contrast to the second method, which utilizes professional role-playing platforms, the third method involves role-playing tasks using general LLMs such as GPT-3.5-Turbo and GPT-4-Turbo, collecting data through interactions with users. While this approach is more efficient in data collection, it may encounter limitations in the role-playing capabilities of general LLMs. Therefore, we will focus more on examining the consistency of the roles in the dialogues collected through this method in later stages.
    \item The fourth method, a fully automatic approach, involves prompting GPT-4-Turbo to engage in self-dialogue by role-playing as both the user and the role-playing agent. This is the most efficient form of collecting dialogue data, leveraging the autonomous capability of general LLMs to simultaneously play the roles of users and role-playing agents in generating dialogue data.
\end{itemize}
   
The prompts for role-playing tasks and automatic self-dialogue generation are provided in Table \ref{tab:prompts-role-playing} and Table \ref{tab:prompts-self-generate}. For the dimension of long-term conversation memory, we construct lengthy dialogue contexts to increase complexity, thereby testing the agent's memory capacity in longer conversational contexts. We achieve this by inserting several rounds of unrelated dialogue between questions and context answers, while ensuring that the unrelated context remains consistent with the current role-playing agent's persona. This approach allows us to extend the dialogue rounds to any length. Prompts for constructing the inserted dialogue context are provided in Table \ref{tab:prompts-inserted}. 
For generating group conversations, the format extends naturally from one-on-one dialogues between users and role-playing agents. In a group setting, members can consist of multiple users interacting with a single role-playing agent, multiple role-playing agents engaging with a single user, multiple users interacting with multiple role-playing agents. Our primary focus lies on scenarios involving multiple role-playing agents. We employ general LLMs such as GPT-4-Turbo to act as different role-playing agents and generate dialogues between their social interactions. Prompts for automatically generating group conversations can be found in Table \ref{tab:prompts-group-dialogue}.

\begin{table*}[t]
    \centering
    \renewcommand{\arraystretch}{1.2}
    \vspace{-0mm}
    \begin{adjustbox}{width=0.95\textwidth}
        \begin{tabularx}{0.95\textwidth}{X}
            \hline
            \hline
            \textbf{Prompt for Role-Playing Tasks} \\
            \hline
            Role Profile: \newline
\{role\_profile\}
\newline
\newline 
You are playing a role-playing game, and your character is \{role\_name\}. \newline
Please adhere to the given profile in terms of character memory, knowledge, and style. You will engage in dialogue with users, following the behavior style of \{role\_name\}. If you understand, please respond with "I understand."
\\
\hline

    \end{tabularx}
    \end{adjustbox}    
    \caption{\label{tab:prompts-role-playing} Prompt for role-playing tasks with GPT-4-Turbo.}
\end{table*}

\begin{table*}[t]
    \centering
    \renewcommand{\arraystretch}{1.2}
    % \vspace{-8mm}
    \begin{adjustbox}{width=0.95\textwidth}
        \begin{tabularx}{0.95\textwidth}{X}
        \hline
            \hline
            \textbf{Prompt for Automatic Self-Dialogue Generation} \\
            \hline
Role Profile: \newline
\{role\_profile\}
\newline
\newline
Example Dialogue: \newline
User: \{user\_utterance\_1\} \newline
Assistant \{assistant\_utterance\_1\} \newline
User: \{user\_utterance\_2\} \newline
Assistant \{assistant\_utterance\_2\} \newline
...... \newline

Please follow the given dialogue example, adhere to the provided profile of \{role\_name\}, 
generate multi-turns conversations between the User and the Assistant (\{role\_name\}). 
The more dialogue turns (For example 30 turns) are better. \newline
The conversations between User and Assistant should follow the format of the given example. 
Dialogue Topic: \{dialogue\_topic\} : \\
\hline

    \end{tabularx}
    \end{adjustbox}    
    \caption{\label{tab:prompts-self-generate} Prompt for automatic self-dialogue generation.}
\end{table*}

\begin{table*}[t]
    \centering
    \renewcommand{\arraystretch}{1.2}
    %\vspace{-8mm}
    \begin{adjustbox}{width=0.95\textwidth}
        \begin{tabularx}{0.95\textwidth}{X}
            \hline
            \hline
            \textbf{Prompts for Constructing Inserted Dialogue} \\
            \hline
            Role Profile: \newline
\{role\_profile\}
\newline
\newline
Previous Dialogue: \newline
...... \newline
Assistant \{assistant\_utterance\} \newline
User: \{user\_utterance\} \newline
\newline
Please follow the provided profile of \{role\_name\}, generate multi-turns conversations between the User and the Assistant. \newline
The generated dialogue should be unrelated to the previously given dialogue content, ensuring diverse and realistic conversation topics while adhering to persona of \{role\_name\}. \\

\hline

        \end{tabularx}
    \end{adjustbox}    
    \caption{\label{tab:prompts-inserted} Prompts for constructing inserted dialogue.}
\end{table*}

\begin{table*}[t]
    \centering
    \renewcommand{\arraystretch}{1.2}
    \vspace{-8mm}
    \begin{adjustbox}{width=0.95\textwidth}
        \begin{tabularx}{0.95\textwidth}{X}
            \hline
            \hline
            \textbf{Prompt for Group Dialogue Generation} \\
            \hline
Profile of \{role\_name\_a\}: 
\newline
\{role\_name\_a\_profile\} 
\newline

Profile of \{role\_name\_b\}: 
\newline
\{role\_name\_b\_profile\} 
\newline

Profile of \{role\_name\_c\}: 
\newline
\{role\_name\_c\_profile\} 
\newline
......
\newline

Example Dialogue: \newline
\{role\_name\_a\}: \{role\_name\_a\_utterance\_1\} \newline
\{role\_name\_b\}: \{role\_name\_b\_utterance\_1\} \newline
\{role\_name\_c\}: \{role\_name\_c\_utterance\_1\} \newline
...... \newline
\{role\_name\_a\}: \{role\_name\_a\_utterance\_n\} \newline
\{role\_name\_b\}: \{role\_name\_b\_utterance\_n\} \newline
\{role\_name\_c\}: \{role\_name\_c\_utterance\_n\} \newline
\newline 
Follow the Dialogue Format, generate multi-turn dialogue between \{role\_name\_a\} and \{role\_name\_b\} and \{role\_name\_c\} ...... \newline
Ensure that each character adheres to their respective personality. The order of dialogue participants can be altered. Aim for as many dialogue turns as possible. \newline
Dialogue scene description: \{dialogue\_topic\}
\\
\hline

        \end{tabularx}
    \end{adjustbox}    
    \caption{\label{tab:prompts-group-dialogue} Prompts for group dialogue generation.}
\end{table*}

\begin{figure*}
    \centering
    % \vspace{-8mm}
    \begin{adjustbox}{width=0.95\textwidth}        \includegraphics{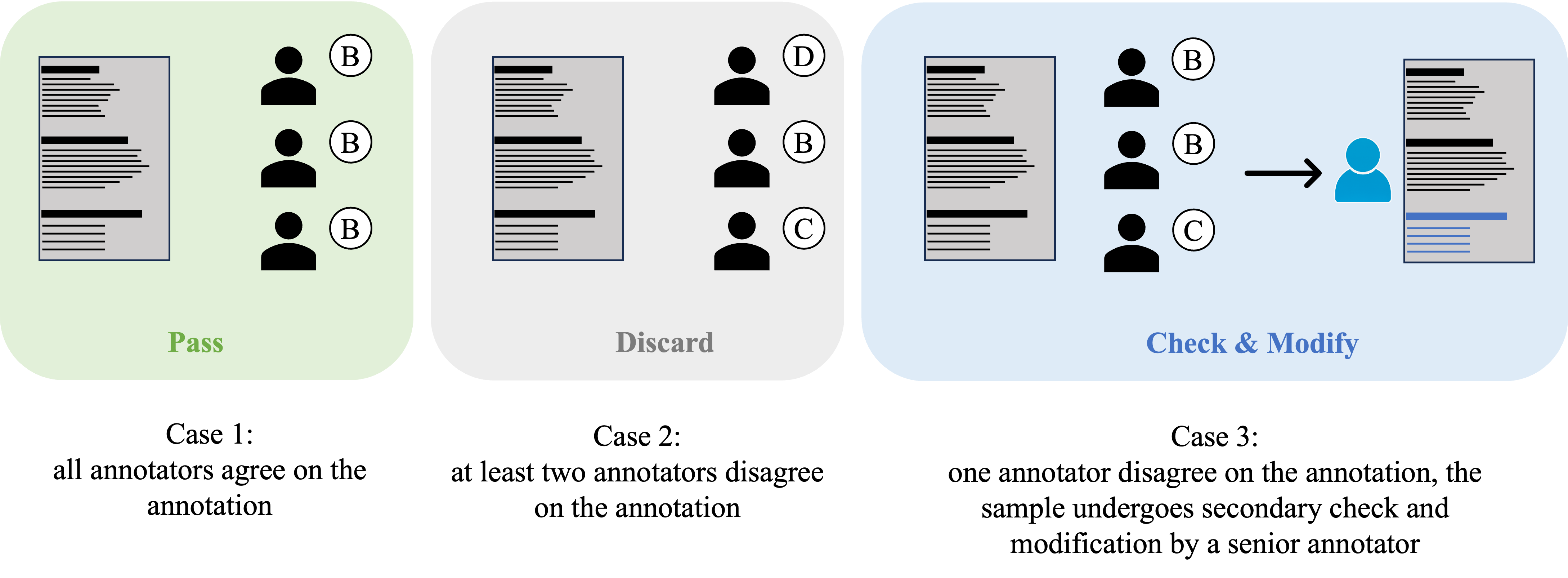}
    \end{adjustbox}
    \caption{Human annotation process.}
    \label{fig:human_annotation_process}
\end{figure*}

\subsection{Question Design}
\label{sec:question-design}

\textbf{For self-awareness:} This includes two subcategories: self-awareness on role style (SA Style) and self-awareness on role knowledge (SA Know.). 
For SA Style, we analyze the corresponding speaking style of a character based on their profile, such as "warm," indicating that the character's speaking style is enthusiastic and cheerful. Since the dialogues constructed in the previous step already adhere to the character's speaking style, we can directly use utterances from the dialogue as correct answers. Additionally, to create negative options, we generate replies with different styles (e.g., "cold," "impersonal"), indicating that these speaking styles do not align with the current character's style setting. It is worth noting that while the speaking style changes, we ensure that the replies still adhere to the contextual coherence. For SA Know., we identify utterances containing character-related knowledge from the dialogue as correct options. For example, some entity information like "Where were you born?" or "Where is your hometown?" This type of information typically follows the character's original setting. We require role-playing agents to possess relevant knowledge when portraying specific characters. Negative options are obtained by modifying entity information in the correct answers.
    
\textbf{For emotional perception:} We construct questions related to situational understanding (EP Situ.) and emotion detection (EP Emo.) based on professional exam questions and relevant open-source datasets \citep{chen2022CPED,hsu2018emotionlines,dilbert,gong-etal-2020-design}.
For EP Situ., we gather exam questions related to situational understanding in psychological counseling scenarios. We filter these questions to exclude those with strong psychological expertise to ensure the assessment focuses on agents' general abilities. We manually collect Level 2 and Level 3 psychological counselor exams, excluding questions on psychology-specific knowledge, while retaining those related to situational and causal understanding.
For EP Emo., we construct emotion understanding data based on open-source datasets and websites. These questions primarily involve agents understanding the psychological states of speakers and interpreting emotions in dialogue. For example, when a speaker says "I hate you," agents need to determine the emotion of this statement based on the context, whether it's hate, like, neutral, etc. We further focus on advanced emotional understanding abilities such as humor and irony. Humor data are collected from websites and the DilBERT dataset \citep{dilbert}, with non-humorous texts used as negative options. For irony emotion understanding, we utilize binary classification data from the Chinese open-source dataset \citep{gong-etal-2020-design} to construct multi-polarity data, selecting one for organization, with the other three non-ironic instances used as negative options.
    
\textbf{For conversation memory:} 
This category includes two subcategories: short-term conversation memory (CM Short) and long-term conversation memory (CM Long).
In SocialBench, questions for other dimensions are presented in multiple-choice format. However, to increase the difficulty of the conversation memory dimension, we utilize an open-domain generation combined with keyword matching approach for this dimension. For example, if an agent previously answered that they had a sandwich for breakfast, after several rounds of conversation, if the user asks again what the agent had for breakfast, we require the agent's response to include the keyword "sandwich." If the agent responds that they had bread for breakfast, since the keyword does not match, we consider the agent unable to correctly recall their previous dialogue content.
For CM Short, we prompt the agent to recall keywords discussed within 40 utterances, while for CM Long, we prompt the agent to recall keywords discussed over 40 utterances.
We evaluate how many of these keywords are recalled.

\textbf{For social preference:} We design questions for three social behavior preferences: positive (Pos.), neutral (Neu.), and negative (Neg.). Group dialogues typically consist of social interactions involving 2 to 10 characters.
We analyze the social preference of a character, and identify behaviors aligning with its preference in the dialogues as correct answers. 
For example, members with a positive social preference tend to engage in behaviors beneficial to the group, such as encouraging teamwork or mediating conflicts within the group. Members with a neutral social preference tend to adopt neutral behaviors within the group, such as aligning with the majority opinion or maintaining a neutral stance in conflicting viewpoints. Conversely, members with a negative social preference tend to engage in behaviors detrimental to the group, such as criticizing others' viewpoints or engaging in competition and arguments with group members.
We analyze the social preference of each character to design negative options. Behaviors contradicting its social preference serve as negative options. For instance, for a character inclined towards teamwork, we would construct exclusionary behaviors as negative options.

\subsection{Dataset Validation}
\label{sec:human-annotation-process}
For multiple-choice questions, we invite three different annotators to label each question. If all three annotators deem the question valid and agree on the answer, it is considered valid.  
As shown in Figure \ref{fig:human_annotation_process}, if all annotators agree on the annotation, it will be selected; if at least two annotators disagree on the annotation, it will be discarded; if only one annotator disagree on the annotation, the question undergoes secondary check by the fourth annotation, it will be modified then selected or be discarded directly. 
For open-domain generation questions, we verify the correctness and validity of keywords provided. 

% \textbf{Profile Verification:} We assess personality contradictions and knowledge hallucinations in profiles to ensure character accuracy. We manually review and modify any erroneous descriptions in profiles, while also ensuring the exclusion of specific personal information such as phone numbers and home addresses.

% \textbf{Dialogue Verification:} Our focus is on ensuring dialogues adhere to principles of \textit{dialogue fluency} and \textit{character fidelity}. For fluency, we manually inspect dialogues for contextual coherence and natural expression. For fidelity, we analyze the speaker's profile to verify if the utterance aligns with the character's speaking style and behavior. Dialogues that do not meet requirements undergo manual correction.

% \textbf{Question Verification:} For multiple-choice questions, we invite three different annotators to label each question. If all three annotators deem the question valid and agree on the answer, it is considered valid. For open-domain generation questions, we verify the correctness and validity of keywords provided. Invalid questions are either modified by experts or discarded. 

For annotators recruiting, we recruit annotators from crowdsourcing companies, and the annotation wages are evaluated and confirmed by the crowdsourcing company. The annotators mainly consist of undergraduate students.

%These datasets can be categorized based on evaluation methods. Firstly, there is the category of open-domain generation combined with human evaluation \citep{shea2023building}. However, this approach faces challenges related to reproducibility and incurs high costs due to the manual evaluation process. there is evaluation through open-domain generation combined with OpenAI API assessment \citep{wang2023rolellm,li2023chatharuhi}. This method may suffer from inaccuracies as GPT-4-Turbo evaluations are not always definitive. And there has been exploration into automated evaluation mechanisms, such as using Rouge-L \citep{wang2023rolellm}, which is not a proper metric. Concurrent work \citep{roleeval,tu2024charactereval,cui2023machine} have begun to explore automated metrics. However, they may have limitations in terms of narrow or simplistic dimensions, and they often do not consider evaluation dimensions specific to persona agents. In contrast, our dataset stands out as it introduces the first-ever evaluation dataset for the sociability of personalized agents.

\begin{table*}[t]
    \centering
    \begin{adjustbox}{width=0.95\textwidth}
        \begin{tabular}{ccc|ccc|ccc}
            \toprule
            \multicolumn{3}{c|}{\textbf{Positive Traits}} & \multicolumn{3}{c|}{\textbf{Neutral Traits}} & \multicolumn{3}{c}{\textbf{Negative Traits}} \\
            Adventurous & Articulate & Attractive & Absentminded & Aggressive & Amusing & Abrasive & Aloof & Angry \\
            Calm & Caring & Cheerful &
            Complex & Conservative & Contradictory &
            Argumentative & Arrogant & Impersonal \\
            Confident & Courageous & Curious & Emotional & Formal & Neutral & Barbaric & Blunt & Childish \\
            Elegant & Humble & Humorous & Mystical & Ordinary & Old-fashioned & Cowardly & Cruel & Fatalistic \\
            Kind & Logical & Optimistic & Stylish & Tough & Whimsical & Gloomy & Lazy & Shy \\
            Passionate & Warm & Witty & Questioning & Sensual & Dry & Envious & Hostile & Melancholic \\ 
            \bottomrule
        \end{tabular}
    \end{adjustbox}
    \caption{Personality traits in SocialBench.}
    \label{tab:personality-traits}
\end{table*}

\section{Evaluation Metrics}
\label{sec:metrics}

For multiple-answer questions, we calculate the accuracy ($Acc_{\text{multiple}}$) using the following formula:

\begin{equation}
    \small
    Acc_{\text{multiple}}=\sum^N_i\frac{\text{Score}_i}{\text{MaxScore}_i},
\end{equation}
where $N$ is the total number of multiple-answer questions. $\text{Score}_i$ is the score obtained for the $i$th question, considering both correct and partially correct options chosen. $\text{MaxScore}_i$ is the maximum achievable score for the $i$th question. For example, if the answer to question i is A, B, then $\text{MaxScore}_i$ is 2. If only A is selected, then $\text{Score}_i$ is 1; if the model selects A, C, and since C is not among A and B, even if A is correct, $\text{Score}_i$ remains 0.

\section{Dataset Statistic}

SocialBench covers 500 characters and 6,000 question prompts involved in 1,000 conversation scenarios and 30,800 multi-turn role-playing utterances. 

\subsection{Personality Traits}
\label{sec:character-types}

% \begin{figure}[ht]
%     \centering
%     \begin{adjustbox}{width=0.48\textwidth}
%         \includegraphics{figures/character-types.pdf}
%     \end{adjustbox}
%     \caption{Character types in SocialBench.}
%     \label{fig:character-types}
% \end{figure}

We follow the definition of personality traits in \citet{ideonomy} to construct profiles, ensuring diversity and comprehensiveness in SocialBench. From the collection of 638 personality descriptors created by \citet{ideonomy}, we selected a subset of easily understandable terms for construction. These selected terms can be categorized into positive, neutral, and negative traits, as illustrated in Table \ref{tab:personality-traits}.

% \subsection{Individual Level and Group Level}SocialBench consists of two main dimensions: individual level and group level. The individual level comprises six sub-dimensions, while the group level comprises three sub-dimensions, as shown in Table \ref{tab:statistic}. Individual level can be split into self-awareness on role style (SA Style), self-awareness on role knowledge (SA Know.), emotional perception on situation (EP Situ.), emotional perception on dialogue emotion (EP Emo.), short-term conversation memory (CM Short), and long-term conversation memory (CM Long). Group level is split into positive (Pos.), neutral (Neu.) and negative (Neg.). 

\section{Data Utilization and Terms of Use}

We utilized the open-source datasets \citep{chen2022CPED,hsu2018emotionlines,dilbert,gong-etal-2020-design}, with their terms of use specifying research purposes only. Similarly, we employed the weights of open-source models and the APIs of closed-source models, strictly adhering to their respective usage agreements for research purposes. Regarding our dataset, it is also restricted to research purposes. We conducted thorough manual checks to ensure the absence of security and offensive issues, particularly sensitive personal information such as phone numbers and home addresses.

\end{document}